\documentclass{article}

\title{Adapting Bandit Algorithms for Settings with Sequentially Available Arms}
\author{Marco Gabrielli, Francesco Trov\`o, Manuela Antonelli\\
	Politecnico di Milano\\
	\{marco.gabrielli, francesco1.trovo, manuela.antonelli\}@polimi.it}
\date{}
\usepackage{amsmath}
\usepackage{amsfonts}
\usepackage{bbm}

\usepackage{algorithm}
\usepackage{algpseudocode}

\usepackage{amsthm}
\newtheorem*{remark}{Remark}

\providecommand{\keywords}[1]{\textbf{Keywords:} #1.}

\usepackage{graphicx}
\graphicspath{ {./plots/} }
\usepackage{color}

\usepackage{placeins}

\newcommand{\alg}{Seq}

\usepackage{graphicx}
\graphicspath{ {./plots/} }
\usepackage{color}

\usepackage{placeins}

\usepackage{thmtools}
\usepackage{thm-restate}

\begin{document}

	\maketitle
	
	\begin{abstract}
	Although the classical version of the Multi-Armed Bandits (MAB) framework has been applied successfully to several practical problems, in many real-world applications, the possible actions are not presented to the learner simultaneously, such as in the Internet campaign management and environmental monitoring settings.
	Instead, in such applications, a set of options is presented sequentially to the learner within a time span, and this process is repeated throughout a time horizon.
	At each time, the learner is asked whether to select the proposed option or not.
	We define this scenario as the \emph{Sequential Pull/No-pull Bandit} setting, and we propose a meta-algorithm, namely Sequential Pull/No-pull for MAB (\alg{}), to adapt any classical MAB policy to better suit this setting for both the regret minimization and best-arm identification problems. 
	By allowing the selection of multiple arms within a round, the proposed meta-algorithm gathers more information, especially in the first rounds, characterized by a high uncertainty in the arms estimate value.
	At the same time, the adapted algorithms provide the same theoretical guarantees as the classical policy employed.
	The \alg{} meta-algorithm was extensively tested and compared with classical MAB policies on synthetic and real-world datasets from advertising and environmental monitoring applications, highlighting its good empirical performances.
\end{abstract}

	\section{Introduction}
In the classical sequential decision-making framework, a learner is presented at each time with a finite set of available options over a finite time horizon, and she/he is asked to select one of them to maximize a specific objective.
For this purpose, a wide range of algorithms have been designed in the Multi-Armed Bandit (MAB) field~\cite{bubeck2012regret}, either resorting to the frequentist~\cite{auer2002finite,audibert2010best,garivier2011klucb} or Bayesian~\cite{kaufmann2012thompson,agrawal2013further} approach.
The adoption of such techniques has been revealed to be effective in a wide range of practical problems, from recommendation systems~\cite{kawale2015efficient} to online advertising~\cite{nuara2018combinatorial}, from networking~\cite{maghsudi2014channel} to dynamic pricing~\cite{trovo2018improving}.
Nonetheless, in many real-world applications, the overall available options are presented sequentially to the learner within a time span, e.g., a day, and are repeated throughout the time horizon, e.g., months.
The learner's decision at a specific time consists of either selecting the single proposed option or the refusal of choosing it for the current time.
We refer to this scenario as the \emph{Sequential Pull/No-pull Bandit (SPNB)} setting.
For instance, an Internet campaign manager faces this type of problem when she/he has to allocate the advertising budget over the day.
In this setting, the advertiser divides the day into a finite number of time slots, representing the time steps of our sequential decision-making process, and sequentially chooses if it is worth allocating some advertising budget to that time slot or not.
The objective is to allocate the budget to the single time slots providing the largest revenue, e.g., clicks or conversions, while minimizing the loss incurred due to the learning process.
The goal of minimizing such a loss is commonly called the Regret Minimization (RM) task in the MAB field.
The environmental quality monitoring, e.g., air or water flow, constitutes another interesting setting that can be modeled as an SPNB setting.
Indeed, an environmental process needs to be monitored to identify the time of the day during which the most critical condition, e.g., in terms of pollution concentration, occurs.
At each time, e.g., hours, during the day, the learner chooses if she/he wants to perform a measurement or not.
Unlike the advertising setting, in this application, the objective is to determine with the highest probability the time of the day, i.e., an hour or a day, during which the highest pollutants concentrations occur.
Instead, the task to identify the best arm with the largest possible confidence is commonly addressed as the Best-Arm Identification (BAI) problem.
The use of MAB-assisted sampling strategies as depicted in this work can lower monitoring costs, removing one of the main hurdles to more widespread use of monitoring campaigns involving advances analytical approaches~\cite{Besmer2017sampling}, which will enable more timely detection of contamination events, reducing their impact on the environment and human health.
Other examples of the SPNB setting are the selection of the optimal time for goods delivery to minimize the time required~\cite{ulku2012optimal}, and the task of detecting the presence of deteriorated equipment in industrial processes~\cite{fouche2019scaling}.
To the best of our knowledge, the design of specifically-crafted MAB algorithms able to exploit the temporal dependency offered by such settings is not known in the literature.

\subsubsection*{Novel Contributions}
In this work, we design a meta-algorithm, namely Sequential Pull/No-pull for MAB (\alg), to improve the performance of classical bandit algorithms in the SPNB setting, exploiting the temporal ordering of the arms present in this scenario.
More specifically:
\begin{itemize}
	\item we cast the SPNB problem in the Multi-Armed Bandit framework, for both the RM and BAI tasks;
	\item we design a framework, namely \alg, to transform any classical MAB algorithm into one for the SPNB setting;
	\item we show that \alg{} has $\mathcal{O}(\log(T))$ regret, $T$ being the time horizon of the learning process, when applied to classical MAB algorithms designed for RM, thus, maintaining the guarantees of such algorithms also in this setting;
	\item we show that applying the \alg{} to a generic classical BAI algorithm still provides the same guarantees;
	\item we provide extensive experimental analysis on both synthetically generated and real-world data coming from an advertisement management problem and a water contaminant monitoring problem to compare the performance of state-of-the-art algorithms with the ones provided by the \alg{} framework. 
\end{itemize}

	\section{Related Works}

The possibility to select multiple arms during each round is usually tackled by Multiple Plays MAB (MP-MAB)~\cite{bubeck2013multiple,Komiyama2015mpts} or Combinatorial MABs (CMABs)~\cite{Chen2013cucb}.
While such approaches generalize the traditional framework by allowing the selection of multiple arms per round, they differ from the one presented here as the number of pulled arms during each time step is constant and given.
Similar to the presented settings, in the Scaling MAB (S-MAB)~\cite{fouche2019scaling} a learner is allowed to pull a variable number of arms between rounds through the use of a scaling policy.
However, differently from the presented work, the number of pulled arms in S-MAB depends on the satisfaction of an efficiency constrain, while in our case this is due to the upper confidence bounds of the arms estimated rewards.
Furthermore, in MP-MABs, C-MABs, and S-MABs, the learner chooses the arms at the beginning of each round before performing any action during that round.
The above-mentioned settings are not directly comparable to ours, as we assume that the learner can choose the pulled arms sequentially, i.e., after discovering the outcome of the previous one, and not at the onset of each round.
Even if specific arms might not be available in the same round when selecting the next arm to pull, e.g., in case the arms are linked to actions at specific times, the presented setting is different from the sleeping bandits one~\cite{Kleinberg2008sleeping} as the eventually unavailable action can be performed in the subsequent round, and the selection is not limited to the arms available in the current round.

From an application point of view, only a few works are present to deal with regret minimization and best arm identification for specific SPNB scenarios.
In the Internet advertising management field, a method to select the most profitable time slot during the day has been presented in~\cite{gasparini2018targeting}.
Nonetheless, this method provides suggestions in an offline fashion, exploiting the information provided from historical data, not including any procedure to include a newly discovered piece of information.
In the environmental monitoring field, a distributed algorithm for detecting contamination events in water systems has been proposed in~\cite{Golovin2010sensors}.
However, such a setting is not comparable to ours, as they assume that obtaining the measurements is less costly than transmitting the results across a network, while we assume that a significant cost is associated with each measurement.
In the considered scenario, the most commonly applied strategies either do not seek to optimize the monitoring strategy, e.g., setting an \textit{a priori} sampling frequency, use external variables as proxy~\cite{Besmer2017sampling} or follow the explore-than-exploit principle~\cite{Gabrielli2021env}. However, these approaches suffer from different drawbacks as proxy variables are not necessarily available in all scenarios, and setting an \textit{a priori} frequency or separating the exploration and exploitation phases is sub-optimal.

	\section{Problem formulation}
In what follows, we defined the Sequential Pull/No-pull Bandit (SPNB) setting.
We assume to have a problem in which a learner is allowed to select among a finite set of $K \in \mathbb{N}$ arms $\{a_1, \ldots, a_K\}$ over a finite time horizon of $T$ time steps.
At each time step $t \in \{1, \ldots, T\}$, the learner is allowed to either select the arm $a_i$ with $i = mod(t, K) + 1$ or decide not to pull it.
We define a round $ro_i$ as a set of $K$ consecutive time steps during which we are presented in a sequence all the available arms, formally $ro_i := \{t_1, \ldots, t_K \in \mathbb{N} \ | \ t_j = \frac{i-1}{K}+j \}$.
During the time horizon $T$ we have a total of $\tau = \lfloor \frac{T}{K}\rfloor$ rounds.\footnote{
	For the sake of simplicity, from now on, we assume that the time horizon $T$ is a multiple of $K$, i.e., $T = \tau K$.}
Each arm $a_i$ at time step $t$ is characterized with a value $x_{i,t}$ of the feedback provided to the learner.
We model the feedback $x_{i,t}$ as a realization of a random variable $X_{i,t}$ drawn from a distribution $\mathcal{D}_i$, whose expected value is $\mu_i := \mathbb{E}[\mathcal{D}_i]$.
As commonly done in the bandit literature, we use Bernoulli distributions to model the feedbacks, i.e., $\mathcal{D}_i \sim Be(\mu_i)$.
We denote with $\mu^* = \max_i \mu_i$ the expected value of the feedback of the optimal arm $a^* = \arg \max_i \mu_i$ and with $X_{*,t}$ the random variable associated with the optimal arm.
An algorithm $\mathfrak{U}$ is a sequential decision-making policy selecting, at each time step $t$, an arm $a_t$ to pull, where the possible options at time $t$ are $a_t = \emptyset$ or $a_t = a_{mod(t, K) + 1}$.
Depending on the setting, an algorithm $\mathfrak{U}$ might have different objectives to optimize: minimize the regret or identify the optimal arm.

\subsection{Regret Minimization}
In the Regret Minimization (RM) framework, the learner's objective is to minimize the loss due to the learning process incurred over time.
More specifically, if the arm to pull for the round is suboptimal, i.e., $a_i \neq a^*$, and the policy $\mathfrak{U}_{RM}$ opts to pull it, the learner gains a reward of $X_{i,t} - X_{*,t}$, i.e., equal to the difference between the values associated to currently considered arm $a_i$ and the optimal one $a^*$.
If the learner decides not to pull such an arm, she/he gets no reward.
Conversely, if the arm to pull at the current time step is the optimal one $a^*$, the learner gains $-X_{*,t}$ reward if she/he opted not to pull it, and a reward equal to $X_{*,t}$ if she/he pulls it.
Formally, the instantaneous reward $Z_t$ is defined as follows:
\begin{equation}
	Z_t := \begin{cases}
		X_{i,t} - X_{*,t} & \text{ if } a_t = a_i \neq a^*\\
		X_{*,t} & \text{ if } a_t = a^* \\
		0 & \text{ if } a_t = \emptyset\\
	\end{cases}.
\end{equation}
The loss incurred by an algorithm $\mathfrak{U}_{RM}$, commonly called \emph{pseudo-regret}, is defined as:
\begin{equation}
	R_T(\mathfrak{U}_{RM}) := \frac{T}{K} {E[Z_t^*]} - \sum_{t=1}^T {E[Z_t]}.
\end{equation}
Notice that, in this setting, the standard definition of pseudo-regret, defined as the difference between the expected value of pulling the optimal arm at a given time step and the expected value of the chosen arm, is not a viable option.
Indeed, using such a pseudo-regret, the na\"ive strategy always opting to pull the available arm at each time step $t$ would suffer null regret.

In what follows, our goal is the design algorithms $\mathfrak{U}_{RM}$ for which the regret $R_T(\mathfrak{U}_{RM})$ grows sub-linearly over time, meaning that the cost per round of the learning process $\frac{R_T(\mathfrak{U}_{RM})}{T} \rightarrow 0$ as $T \rightarrow + \infty$.

\subsection{Best-Arm Identification}

In the Best-Arm Identification (BAI) framework, the learner's objective is to identify the arm providing the largest expected reward, minimizing the chances of an error in the identification.
In this setting, we require to provide both an algorithm $\mathfrak{U}_{BAI}$, a.k.a.~sampling strategy, a stopping rule $\mathfrak{S}$, providing the learner a time $t$ at which the algorithm has finished the process, and a procedure to provide a final guess $\mathfrak{G}$, providing a guess $\hat{a}^*_t$ at time $t$ on the optimal arm.
In this setting, we would like to provide PAC guarantees for a given tuple $(\mathfrak{U}_{BAI}, \mathfrak{S}, \mathfrak{G})$ that:
\begin{equation}
	\mathbb{P}(\hat{a}^*_t \neq a^*) \leq \delta_t,
\end{equation}
where $\delta_t \in (0, 1)$ is a confidence level.
Depending if either we want to fix in advance the stopping time of the algorithm $t$ or the confidence $\delta_t$ of the algorithm, we are in the so called fixed-budget setting, or the fixed-confidence one, respectively.
See~\cite{audibert2010best} for more details.

\begin{remark} \label{rem:classical}
	The first idea to apply standard MAB techniques on the available $K$ arms consists of selecting a single arm at each round $ro_i$.
	A pseudo-code describing such a na\"ive approach is provided by Algorithm~\ref{alg:mab_generic} present in Appendix~\ref{app:algo}.
	Therefore, the standard MAB setting is played over a time horizon of $\tau = \frac{T}{K}$ time steps since we are allowed to pull exactly one arm per round.
	This constitutes a suboptimal strategy since we are missing the chance of selecting multiple arms to gather more information, especially in the first rounds in which the estimated values for the arms are uncertain, and we might want to perform a larger number of exploratory pulls per round.
\end{remark}

\begin{remark} \label{rem:roundr}
	A specific class of MAB algorithms commonly referred to as elimination algorithms allows a different approach for the SPNB setting for both the RM and BAI settings.
	Indeed, these algorithms iteratively exclude one or more arms that are likely not to be optimal during the learning process, and since the arms selection process occurs in a round-robin fashion, they can be applied to the setting so that at each round $ro_i$ they can pull at most $K$ arms.
	Therefore, slightly modifying their definition might provide significant improvement in the SPNB setting.
\end{remark}

In the following sections, we will analyze the theoretical guarantees and the empirical performance of the above-mentioned approaches and the proposed \alg{} framework, crafted explicitly for the SPNB setting.
	\section{The Sequential Pull/No-pull MAB Algorithm}

In what follows, we propose a meta-algorithm applicable to any classical MAB algorithm, either for the RM or BAI tasks, which is better suited to the SPNB setting.
The founding idea is that we should pull an arm $a_i$ any time we are allowed to do that, and a classical MAB algorithm would pull it, instead of waiting for the next phase $ro_i$ to pull it as a classical MAB algorithm would do.
The pseudo-code of the proposed approach, namely Sequential Pull/No-pull MAB (\alg) is presented in Algorithm~\ref{alg:seqmab}.
It requires as input a classical MAB policy $\mathfrak{U}_{MAB}$, either for RM or BAI, and an ordered set of arms to choose from $\{ a_1, \ldots, a_K \}$.
At first, it initializes the policy $\mathfrak{U}_{MAB}$ and a counter for the current number of pulls $n$.\footnote{
If the policy requires a number of steps for the initialization, we should remove them from the main loop and perform such a procedure at this step.
In this case, we should also increase the counter $n$ of a number of rounds corresponding to the ones used for the initialization.}
At each time step $t$, we run the MAB algorithm $\mathfrak{U}_{MAB}$ as if we had a total number of pulls equal to $n$.
This provides the next arm to pull $a_{MAB}$ according to $\mathfrak{U}_{MAB}$.
If this arm is the one we are allowed to pull for this round, i.e., $a_{MAB} = a_{mod(t,K)+1}$ we pull the arm, and collect the feedback $x_{MAB,t}$ from the selected arm $a_{MAB}$.
Otherwise, we opt not to pull anything and proceed to the next round $t + 1$ without any update.
Differently from the na\"ive application of $\mathfrak{U}_{MAB}$ described in the previous section, the \alg{} approach allows to perform multiple pulls per round if this is advised by the strategy $\mathfrak{U}_{MAB}$.

\begin{algorithm}
	\caption{\alg($\mathfrak{U}_{MAB}$)}
	\label{alg:seqmab}
	\begin{algorithmic}[1]
		\State \textbf{Input:} MAB algorithm $\mathfrak{U}_{MAB}$, arm set $\{ a_1, \ldots, a_K \}$, time horizon $T$
		\State Initialize $\mathfrak{U}_{MAB}$
		\State $n \gets 0$
		\State $a_{MAB} \gets 1$
		\For{$t \in \{1, \ldots, T\}$}
		\If{$a_{MAB} = a_{mod(t,K)+1}$}
		\State Pull arm $a_{MAB}$
		\State Collect feedback $x_{MAB,t}$
		\State $n \gets n + 1$
		\State Update $\mathfrak{U}_{MAB}$
		\State $a_{MAB} \gets \mathfrak{U}_{MAB}(n)$
		\EndIf
		\EndFor
	\end{algorithmic}
\end{algorithm}

\subsection{Regret Analysis for the Regret Minimization Algorithms}
In this section, we derive the pseudo-regret for the classical algorithms applied directly to the SPNB setting, as presented before in Remark~\ref{rem:classical}, and for the \alg{} framework.

In the case we apply a classical algorithm for RM, e.g., UCB1~\cite{auer2002finite}, Bayes-UCB~\cite{kaufmann12bucb}, or Thompson Sampling~\cite{thompson1933ts}, to the SPNB setting, as specified in Algorithm~\ref{alg:mab_generic} (provided in Appendix~\ref{app:algo}) we have that the pseudo-regret is:

\begin{restatable}[]{mythm}{thmregclassical} \label{thm:classical}
	Using a classical RM algorithm $\mathfrak{U}_{RM}$, with guarantees on the expected number of pulls of the suboptimal arms of $\mathbb{E}[T_i(t)] \leq C_i \log(t) + A_i$, where $C_i$ is $o(1)$ and $A_i$ is $o(\log(t))$, over a time horizon of $t$, on the SPNB setting it suffers a pseudo-regret of:
	\begin{equation}
		R_T(\mathfrak{U}_{RM}) \leq \sum_{a_i \neq a^*} (\mu^* + \Delta_i) [C_i \log(T) + A_i - C_i K],
	\end{equation}
	where $\Delta_i := \mu^* - \mu_i$ is the gap between the expected reward of the optimal arm $a^*$ and a suboptimal arm $a_i$.
\end{restatable}
The full proof of Theorem~\ref{thm:classical}, as well as those of the following theorems, is deferred to Appendix~\ref{app:proof} for space reasons.\footnote{With $f(t) = o(g(t))$ we denote two functions for which as $t \rightarrow \infty$ if for every positive constant $\varepsilon$ there exists a constant $N$ such that $|f(t)| \leq \varepsilon g(x)$ for all  $t \geq N$.}
For the UCB1 algorithm, for which the bound on the expected number of pulls is bounded by the constants $C_i = \frac{8}{\Delta_i^2}$ and $A_i = (1 + \frac{\pi^2}{3})$ (see~\cite{auer2002finite} for details), we have a bound on the regret of:
\begin{equation*}
	R_T(UCB1) \leq \sum_{a_i \neq a^*} \frac{8 (\mu^* + \Delta_i)}{\Delta_i^2} \log(\tau) + \sum_{a_i \neq a^*} \left( 1 + \frac{\pi^2}{3} -  \frac{8 K}{\Delta_i^2} \right) (\mu^* + \Delta_i).
\end{equation*}
Conversely, for the Bayes-UCB algorithm we have that $C_i := \frac{1 + \epsilon}{KL(\mu_i, \mu^*)}$ and $A_i := \frac{c \log(\log(t))}{KL(\mu_i, \mu^*)} + K_c (\log(\log(t)))^2 + o(1)$, for any $\epsilon > 0$, $c > 5$ and $K_c > 0$, providing a bound of:
\begin{equation*}
	R_T(Bayes-UCB) \leq \sum_{a_i \neq a^*} \frac{(1 + \epsilon) (\mu^* + \Delta_i)}{KL(\mu_i, \mu^*)} \log(\tau) + o((\log(\log(t)))^2),
\end{equation*}
where $KL(a, b)$ is the Kullback-Leibler divergence of two Bernoulli variable with expected values $a$ and $b$.
Thanks to the Pinsker's inequality stating that $\frac{1}{KL(\mu_i, \mu^*)} \leq \frac{1}{2 \Delta_i^2}$ the bounds can also be written as:
\begin{equation*}
	R_T(Bayes-UCB) \leq \sum_{a_i \neq a^*} \frac{(1 + \epsilon) (\mu^* + \Delta_i)}{2 \Delta_i^2} \log(\tau) + o((\log(\log(t)))^2).
\end{equation*}
Similarly, for the Thompson Sampling algorithm we have:
\begin{equation*}
	R_T(TS) \leq \sum_{a_i \neq a^*} \frac{(1 + \epsilon) (\mu^* + \Delta_i)}{KL(\mu_i, \mu^*)} \log(\tau) + o((\log(\log(t)))),
\end{equation*}
since $C_i := \frac{1 + \epsilon}{KL(\mu_i, \mu^*)}$ and $A_i := o(\log(\log(t)))$.

As mentioned before, the use of the so called \emph{elimination algorithms} in the SPNB setting, due to their round-robin arm selection approach, allow their application in a more efficient way, and this reflects in a better regret bound.
For instance, the UCBrev algorithm~\cite{Auer2010ucbrev} operates as follows:
pulls all the arms in a round robin fashion until all the arms have a given number of pulls; after that it uses Hoeffding's bounds to exclude those arms which are likely to be suboptimal, and iterates until the total number of pullsa reached the time horizon.
The modification of the UCBrev algorithm which selects multiple arms per round, from now on denoted with UCBrev+, is detailed by Algorithm~\ref{alg:ucbrev} in Appendix~\ref{app:algo}.
Even if the UCBrev+ exploits better than the other RM approaches the temporal dependency in the SPNB setting, a specifically crafted analysis on its regret fails in providing a better regret bound than the one provided in Theorem~\ref{thm:classical}.
See Appendix~\ref{app:proof} for details.

Finally, we show that the use of a generic RM algorithm in the \alg{} framework provides an upper bound on the pseudo-regret of the same order of using a generic RM algorithm in the SPNB setting:
\begin{restatable}[]{mythm}{thmregseq}
	Given a classical RM algorithm $\mathfrak{U}_{RM}$, with guarantees on the expected number of pulls of the suboptimal arms of $\mathbb{E}[T_i(t)] \leq C \log(t) + A$ over a time horizon of $t$, the Seq$(\mathfrak{U}_{RM})$ algorithm on the SPNB setting over a time horizon of $T$ rounds suffers from a pseudo-regret of:
	\begin{equation}
		R_T(\alg(\mathfrak{U}_{RM})) \leq \sum_{a_i \neq a^*} (\Delta_i+\mu^*) [C_i \log(T) + A_i].
	\end{equation}
\end{restatable}
We remark that, even if the design of \alg{} allows to pull an arm at each time step $t$, it suffers from a regret of the same order of the one in Theorem~\ref{thm:classical}.
In the experimental section, we will analyse the empirical improvement of the \alg{} approach.

\subsection{PAC Analysis for the Best-arm Identification Goal}

The focus in the BAI problem is to select with high probability, at the end of an exploration procedure, the optimal arm $a^*$.
In the SPNB scenario, one might apply a generic BAI algorithm in a straightforward way by selecting the arm to pull once for each round $ro_i$, therefore selecting a single arm to pull every $K$ time steps.
This approach is exemplified again by Algorithm~\ref{alg:mab_generic} provided in Appendix~\ref{app:algo}.
This approach has the following guarantees:
\begin{restatable}[]{mythm}{thmbaiclassical} \label{thm:baicla}
	A classical BAI algorithm $\mathfrak{U}_{BAI}$, with guarantees of $\delta_t(\mathfrak{U}_{BAI}) \leq C_1 t K $ on the classical MAB setting, on the SPNB setting, provides a confidence of:
	\begin{equation}
		\delta_t(\mathfrak{U}_{BAI}) \leq C_1 t K^2.
	\end{equation}
\end{restatable}
Notice that depending if we are in the fixed confidence or in the fixed budget BAI setting we set $\delta_t$ and $t$, respectively, and compute the corresponding $t$ and $\delta_t$, respectively.
The additional linear dependence on $K$ w.r.t.~the standard BAI setting is due to the fact that the learner is allowed to pull a single arm at each round $ro_i$, performing a total of $\tau$ pulls, while the potentially available pulls are $t$ in total.
For instance, this results states that if we choose the UCBE algorithm~\cite{audibert2010best}, it provides a guarantee of $\delta_t(UCBE) \leq 2 t K^2 \exp{\frac{2\sum_{i=1}^K 1/\Delta_i^2}{25}}$, and if we choose the SR algorithm~\cite{audibert2010best} with a budget of $\tau$ rounds, we have:
\begin{equation}
	\delta_T(SR) \leq \frac{K(K-1)}{2} \exp \left( - \frac{T - K^2}{K\overline{\log}(K) H_2} \right),
\end{equation}
where $\overline{\log}(K) := \frac{1}{2} + \sum_{i=2}^K \frac{1}{i}$, $H_2 := \max_{ i \in [K]} \frac{i}{\Delta_{(i)}^2}$, and the sequence $\{ \Delta_{(i)} \}_{i=1}^K$ is the ordering of the gaps $\Delta_i$ in increasing order, i.e., formally $\min_{i} \Delta_i = \Delta_{(1)} = \Delta_{(2)} \leq \ldots \leq \Delta_{(K)} = \max_{i} \Delta_i$.

Even in the BAI setting, a slightly different use of an elimination algorithm might be provide some improvement w.r.t.~the above-mentioned approach.
For instance, consider the SR algorithm~\cite{audibert2010best} that works as follows: it divides the total pulls into phases, during which all the available arms are pulled the same number of times, it eliminates a single arm at the end of each phase, and repeats the process until a single arm remains.
This procedure allows to select multiple arms per round.
The definition of the algorithm derived from SR and selecting multiple arms per round, denoted from now on as SR+, is provided by Algorithm~\ref{alg:sr} in Appendix~\ref{app:algo}.
Using the SR+ algorithm in the SPNB setting, we show that:
\begin{restatable}[]{mythm}{thmbaisr}
	The SR+ algorithm with a budget of $n = \frac{(2 T - 1) \overline{\log}(K)}{2K}+K$ on the SPNB setting, provides a confidence of:
	\begin{equation}
		\delta_T(SR+) \leq \frac{K(K-1)}{2} \exp \left( - \frac{2T-1}{2 H_2}, \right).
	\end{equation} 
\end{restatable}
Notice that this result has a better scaling factor of $\approx \frac{K}{\overline{\log}(K)} \geq 2$ w.r.t.~the one obtained by the SR algorithm.
This is due to the fact that this modified version is able to pull multiple arms per round, packing as much as possible the exploratory phases over the time horizon $T$.

Finally, if we apply the \alg{} meta-algorithm to any BAI algorithm, it is trivial to show that the guarantees are the same as the ones provided by a generic BAI algorithms in Theorem~\ref{thm:baicla}, formally:
\begin{restatable}{corollary}{thmbaiseq}
	Consider a classical BAI algorithm $\mathfrak{U}_{BAI}$, with guarantees of $\delta_t(\mathfrak{U}_{BAI}) \leq C_1 t K $ on the classical MAB setting.
	The \alg($\mathfrak{U}_{BAI}$) algorithm, on the SPNB setting, provides a confidence of:
	\begin{equation}
		\delta_t(\alg(\mathfrak{U}_{BAI})) \leq C_1 t K^2.
	\end{equation}
\end{restatable}
Even in the BAI setting, we are not able to provide a tighter result since we have no strong guarantees that this approach is selecting a more than one arm at each round $ro_i$.
Nonetheless, we will show in the next section how this approach is able to provide better empirical performance w.r.t.~the straightforward application of such techniques to the SPNB problem.

	\section{Experimental Results}

We conducted numerical simulations to assess the experimental performance of \alg{} with the ones of classical MAB in for RM and BAI settings, and the newly-introduced UCBrev+ and SR+.\footnote{
	We want to remark that the UCBrev, and SR algorithms were not reported here since their performances were strictly worse than their counterparts UCBrev+, and SR+, respectively.}
We experimented on synthetically generated data and two real-world dataset coming from an advertising and environmental monitoring applications.

\subsection{Performance evaluation metrics}
The performance of all algorithms was evaluated with several metrics in both RM and BAI settings.

\subsubsection{RM settings}
For both synthetic and real-world datasets we evaluate the performance of each RM algorithm $(\mathfrak{U})$ in terms of:
\begin{itemize}
	\item $\hat{R}_T$ the empirical pseudo-regret, formally $\hat{R}_T(\mathfrak{U}) = \frac{T \mu^*}{K} - \sum_{t=1}^T \mu_{i(t)}$, where $i(t)$ is the index of the arm chosen by $\mathfrak{U}$;
	\item \textit{NPR} the number of pulls per round, formally \textit{NPR}$(\mathfrak{U}, ro_i) = \sum_{t \in ro_i} \mathbbm{1} \{a_{i(t)} = a_{mod(t,K)+1} \}$;
	\item $Opt^*$ the percentage of pulls of the optimal arm $a^*$ over the number of pulls, formally $Opt^*(\mathfrak{U}) := \frac{\sum_{t | a^* = a_{mod(t,K)+1}} \mathbbm{1} \{a_{i(t)} = a_{mod(t,K)+1} \}}{\sum_{t} \mathbbm{1} \{a_{i(t)} = a_{mod(t,K)+1} \}}$;
	\item $Opti^*$ the percentage of the rounds for which the algorithm $\mathfrak{U}$ pulled the optimal arm, formally $Opti^*(\mathfrak{U}) := \frac{\sum_{t | a^* = a_{mod(t,K)+1}} \mathbbm{1} \{a_{i(t)} = a_{mod(t,K)+1} \}}{\tau}$.
\end{itemize}

\subsubsection{BAI settings}
For the synthetic datasets the performance of the algorithms was evaluated in terms of:
\begin{itemize}
	\item $\hat{\delta}_t$ the percentage of selecting the a suboptimal arm, formally $\hat{\delta}_t(\mathfrak{U}) := \frac{1}{n}\sum_{e=1}^n \mathbbm{1} \{a^*_t,e \neq a^*\}$;
	\item $\Psi_{\mathfrak{U}, rounds}$ the percentage of rounds used by \alg(UCBE) and SR+ before reaching the stopping criterion compared to UCBE, formally $\Psi_{\mathfrak{U}, rounds} := \frac{\theta_{\mathfrak{U}}-\theta_{UCBE}}{\theta_{UCBE}}$, where $\theta_\mathfrak{U}$ and $\theta_{UCBE}$ are the number of rounds used by a generic algorithm $\mathfrak{U}$ and UCBE, respectively.
	This metric was used to evaluate \alg(UCBE)-LP and SR+, as their number of pulls is equal to UCBE by design;	
	\item $\Psi_{pulls}$ the percentage of pulls used by \alg(UCBE) before reaching the stopping criterion compared to UCBE, formally $\Psi_{pulls} := \frac{t' - \tau}{\tau}$, where $t'$ are the pulls performed by \alg(UCBE).\footnote{Since UCBE pulls a single arm per round, the number of pulls in $\tau$ rounds is $\tau$.} This metric was used to evaluate only \alg(UCBE)-LR;
\end{itemize}
where $n = 100$ is the number of independent runs performed for each scenarios.\\
As the real-world data was used to evaluate the performance of the algorithms in the BAI settings, it was not possible to test multiple realization of the problem. 
For this reason, the algorithms were compared by assessing their correctness, evaluated only through the use of $\hat{\delta}_t$, varying the number of rounds available before selecting the best arm.

\paragraph{Figures}
In the provided figures, solid lines, bars and dots show the estimated mean values, while shaded areas and confidence bars provide the estimated $95\%$ confidence interval of the mean.
Confidence intervals were omitted from the plots where negligible.

\subsection{Synthetic Datasets}
\label{sect:synth-data}

\subsubsection{Regret Minimization}
The performances of the \alg{} approach for RM task was tested comparing bayes UCB (bUCB)~\cite{kaufmann12bucb} and TS~\cite{thompson1933ts} applied in the SPNB setting with their counterparts, i.e., Seq(bUCB) and Seq(TS), respectively.
Moreover, we applied the adapted version of the UCBrev~\cite{Auer2010ucbrev} algorithm, UCBrev+.
We simulated three SPNB scenarios, each of which having a different number of arms $K \in \{10, 25, 50\}$, over a period composed of $\tau = 1,000$ rounds.
The expected arms rewards $\mu_i$ were uniformly sampled in $[0,1]$, conditioned on having $\min_i \Delta_i = 0.1$.
Experimental results are averaged over $100$ independent runs.

\begin{figure}[t!]
	\centering
	\includegraphics[width=\textwidth]{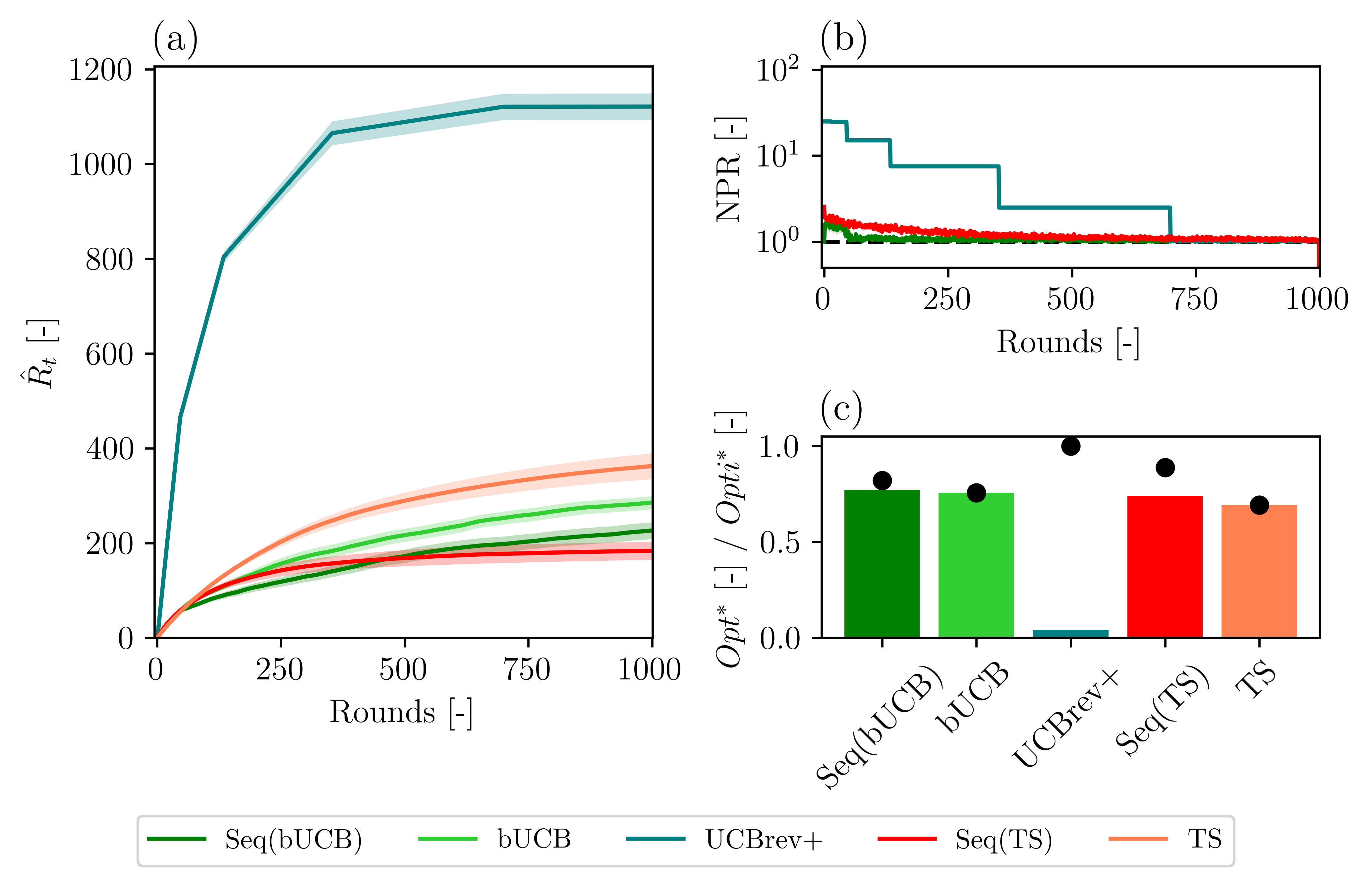}
	\caption{Results for the synthetic setting with $K = 25$ arms: (a) $\hat{R}_T$, (b) NPR, (c) $Opt^*$ (shown as bars) $Opti^*$ (shown as points) for the analysed algorithms.}
	\label{fig:MAB-synth25}
\end{figure}

\paragraph{Results}
The results for the experiments with $K = 25$ are provided in Figure~\ref{fig:MAB-synth25}.\footnote{
The results corresponding to the experiments with $K = 10$ and $K = 50$ are provided in the Supplementary Materials, given that they are in line with those of the case $K = 25$.}
At first, Figure~\ref{fig:MAB-synth25}b confirms that the proposed methodology is capable of increase the number of pulls per round.\footnote{The number of pulls for bUCB and TS are not shown in Figure~\ref{fig:MAB-synth25}b since they are allowed to deterministically pull a single arm per round.}
Indeed, pulling multiple arms per round allows \alg(bUCB) and \alg(TS) to reduce the pseudo-regret suffered over time compared to the traditional counterparts (Figure~\ref{fig:MAB-synth25}a).
Moreover, both algorithms select the optimal arm a number of times larger than its counterpart (see Figure~\ref{fig:MAB-synth25}c). In fact, the larger decrease in pseudo-regret observed for \alg(TS) compared to \alg(bUCB) can be linked by its higher increase of $Opti^*$ from its traditional counter.
On the other hand, UCBrev+ pulled the optimal arm with a large probability ($Opti^*(UCBrev+) \approx 1$) over the rounds.
However, it still achieved significantly worse performances in terms of pseudo-regret (Figure~\ref{fig:MAB-synth25}a).

Looking at Figure~\ref{fig:MAB-synth25}b, we notice that the differences between the \alg{} algorithms and UCBrev+ are due to the dramatically larger \textit{NPR} of the latter algorithm.
The two \alg{} algorithms perform only a limited \textit{NPR} during the first rounds, quickly dropping to a single pull per round.
Conversely, UCBrev+ start pulling a large number of arms, i.e., equal to $K$ in the first rounds, and decrease slower than the other two.

Considering $Opt^*$ in Figure~\ref{fig:MAB-synth25}c, the difference between \alg(TS),  \alg(bUCB) and UCBrev+ is also due to the fact that \alg{} algorithms allocate a percentage of pulls to $a^*$ comparable to the traditional counterparts, while UCBrev+ select $a^*$ with a lower percentage of the total number of pulls. Therefore, UCBrev+ is selecting multiple arms per round, but the large share of suboptimal selections, linked with the slower \textit{NPR} decrease, dramatically increase its regret over time. 

\subsubsection{Best-Arm Identification}

The performance of the Seq framework in the BAI settings was assessed by conducting experiments similar to the one presented by Audibert and coworkers~\cite{audibert2010best} and adapting them to the SPNB setting.
More specifically, the seven SPNB BAI experiments tested were set up considering the arms with expected feedback $\mu_i$ equal to the original experiments, but ordered following the value of the subscript $i$ within each $ro_i$, and setting a stopping criterion based on the round reached by the algorithms.\footnote{The arms $\mu_i$ distribution of the seven experiments are shown in the Supplementary Materials}
We compared the UCBE~\cite{audibert2010best} and SR+ algorithms with Seq(UCBE), the Seq counterpart of UCBE.
Seq(UCBE) and UCBE were tested considering different values of the parameter $c \in \{1, 2, 4, 8\}$.
Each experiment was repeated over $10$ independent runs, each one consisting of 100 problem evaluations.
We designed two different flavors of the Seq(UCBE), depending on the stopping criterion we adopted, which correspond to two different usage cases.
In the Seq(UCBE)-LP algorithm, the best arm is suggested when reaching a number of pulls equal to UCBE.
This scenario reflects the case where a strong limit on the budget is imposed, i.e., the number of pulls is limited, and an earlier identification of the best arm is favored.
Conversely, Seq(UCBE)-LR provide a guess on the optimal arm after an equal number of rounds $\tau$ was given to the two algorithms and is better suited for a situation in which a strong limit on the budget does not exist, but an increase in the algorithm performance is sought.

\begin{figure}[t!]
	\centering
	\includegraphics[width=\textwidth]{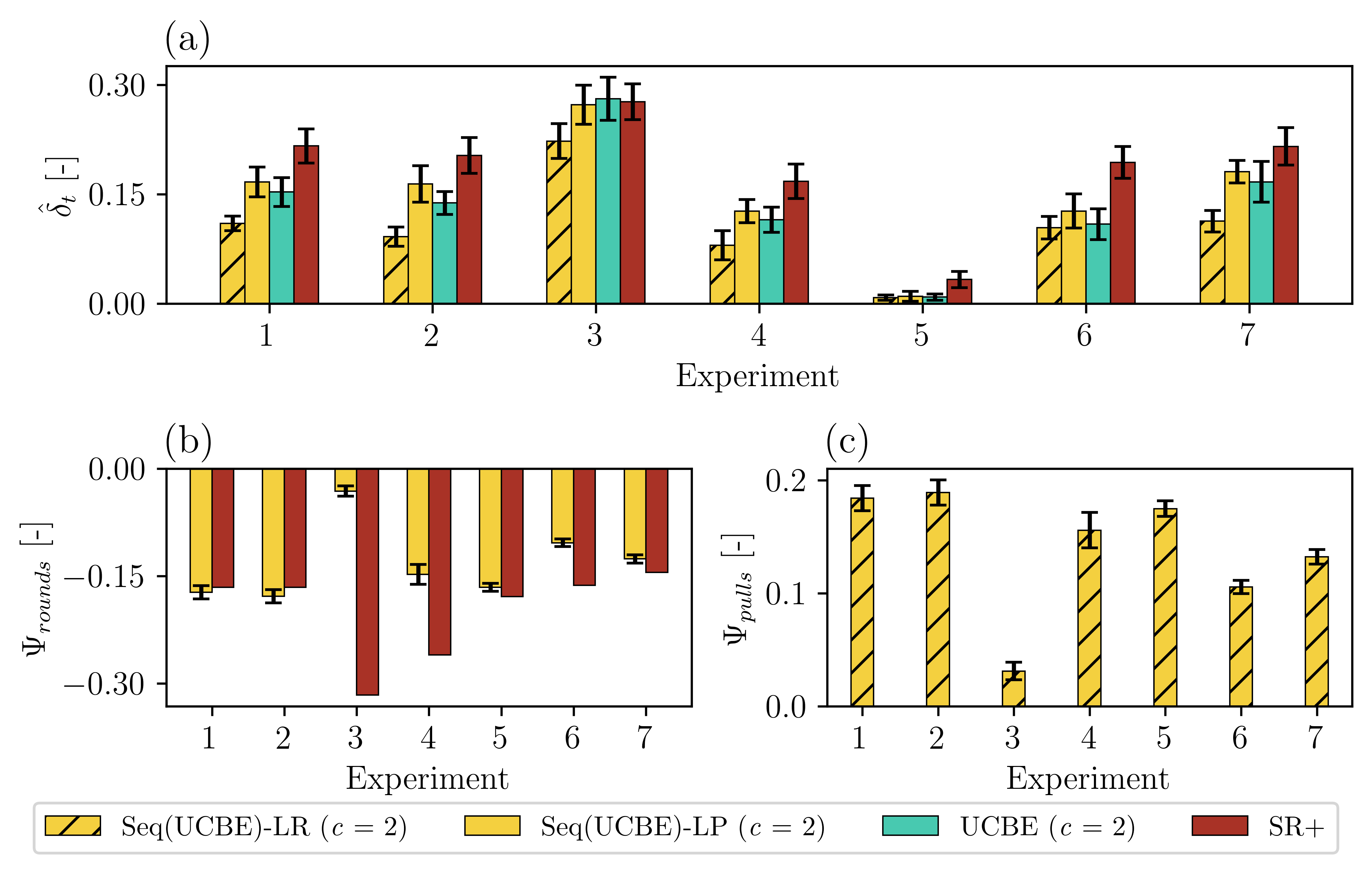}
	\caption{$\hat{\delta}_t$ (a), $\Psi_{rounds}$ (b) and $\Psi_{pulls}$ (c) for both scenarios of Seq(UCBE) (\textit{c} = 2), UCBE (\textit{c} = 2) and SR+.}
	\label{fig:BAI-synth-c2}
\end{figure}

\paragraph{Results}

Figure~\ref{fig:BAI-synth-c2} shows the outcome of the experiments using $c = 2$.\footnote{The results corresponding to other values of $c$ are reported in Appendix~\ref{app:additional} and are in line with those presented in this section.}
The Seq(UCBE)-LP algorithm performs better than SR+ and comparably to UCBE in terms of $\hat{\delta}_t$ (Figure~\ref{fig:BAI-synth-c2}a).
However, the results of Seq(UCBE) were obtained, requiring a significantly lower number of rounds than UCBE.
More specifically, in Figure~\ref{fig:BAI-synth-c2}b, the $\Psi_{\alg(UCBE), rounds}$ is between $-3\%$ and $-18\%$, depending on the experiment, w.r.t.~UCBE. Such reduction is, in the majority of the experiments, similar to what obtainable using SR+. 
The Seq(UCBE)-LR algorithm generically provides a significant reduction in terms of percentage of correct identification $\hat{\delta}_t$ when compared to the UCBE and SR+ ones (Figure~\ref{fig:BAI-synth-c2}a), due to an increased number of pulls performed during the same amount of rounds.
More specifically, the increase in terms of $\Psi_{pulls}$ varies from $3\%$ and $19\%$, depending on the experiment (Figure~\ref{fig:BAI-synth-c2}c).

\subsection{Real-World Datasets}
In this section, we test the proposed algorithms against their traditional counterparts on real-world problems.

\subsubsection{Regret Minimization}
\begin{figure}[t!]
	\centering
	\includegraphics[width=0.75\textwidth]{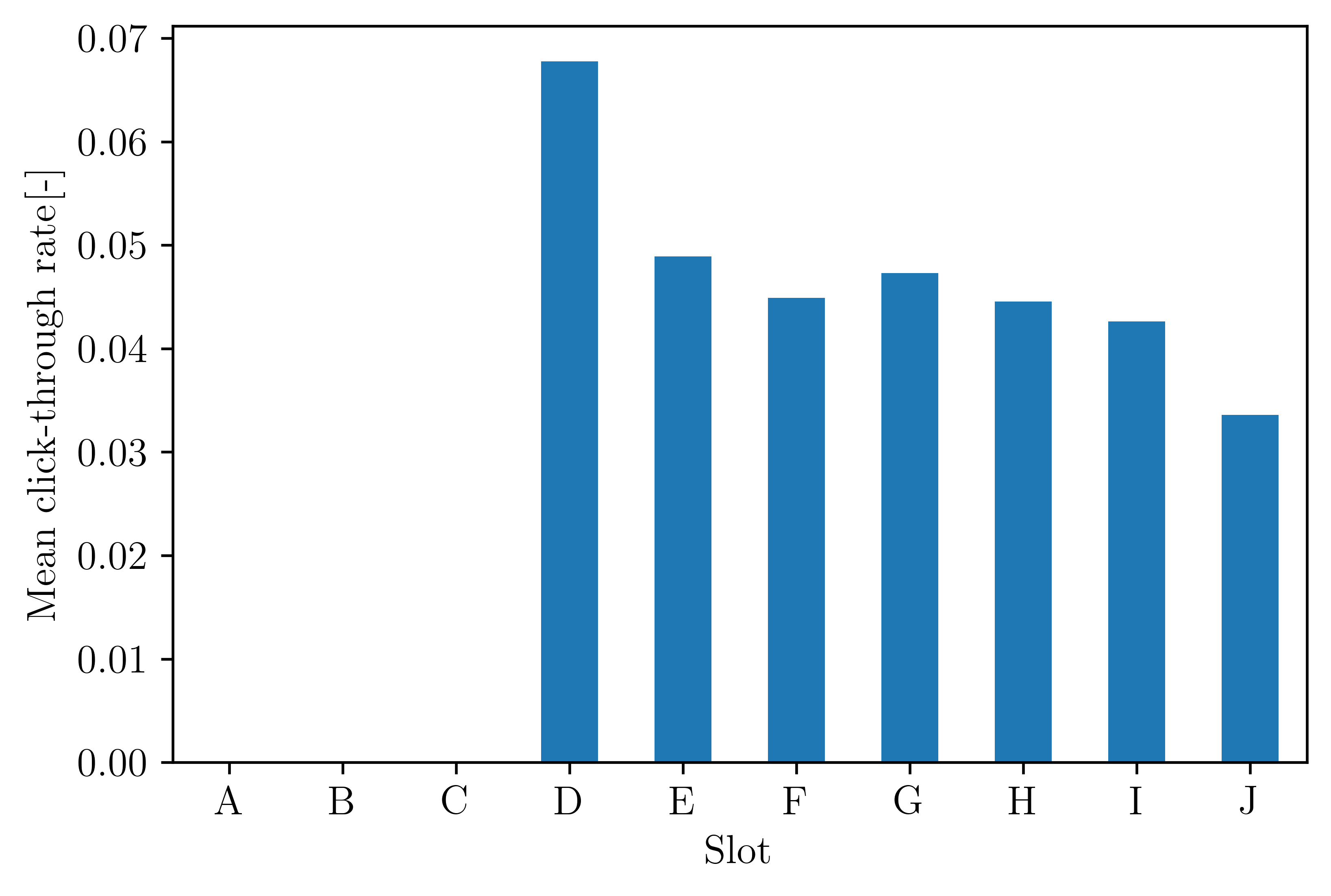}
	\caption{Mean click-through rates calculated for a selected article.}
	\label{fig:CTRmean}
\end{figure}

The performance of the RM algorithms have been tested on the Yahoo! Front Page Today Module User Click Log Dataset~\cite{li2011yahoo}.
Such dataset contains a user click log for the articles displayed in the Featured Tab of the font page of Today Module on Yahoo! of a few days in May $2009$.
Similarly to~\cite{liu2017cd} and~\cite{mellor2013cd}, average click-though rates were calculated from the first day in the dataset by taking the mean of the articles click-through rate every at given intervals.
Such intervals were set approximately equal splitting the day in $10$ slots with an equal number of accesses.
The slots were kept in their order and named following the alphabet letters.
As each article is not displayed throughout the entire day, only the time slots in which the article has been displayed were considering for the RM problem, as depicted in Figure~\ref{fig:CTRmean}.
As in this scenario each round represents one day in which the articles are to be displayed, and the entire experiments has been conducted in a time horizon $T$ of $2$ years.
The obtained click-through rates distributions were used to generate $100$ independent simulations for each article present in the dataset.

\begin{figure}[t!]
	\centering
	\includegraphics[width=\textwidth]{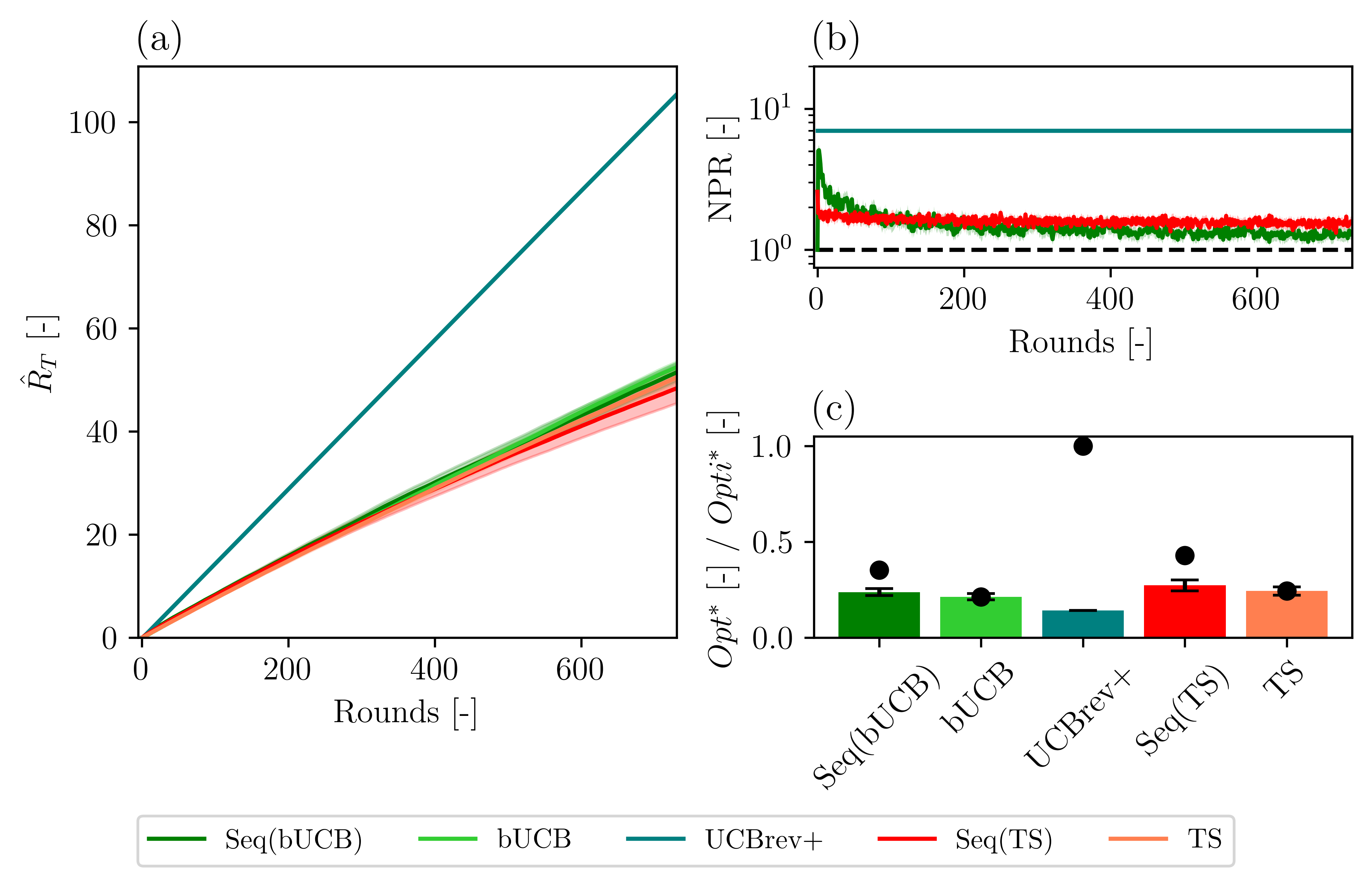}
	\caption{$\hat{R}_T$ (a), NPR (b) and $Opt^*$ (shown as bars) and $Opti^*$ (shown as points) (c) for Seq(bUCB1, bUCB, UCBrev+, Seq(TS), and TS for the article whose arms distribution is shown in Figure \ref{fig:CTRmean}.}
	\label{fig:MAB-rw}
\end{figure}

The empirical results obtained with the real-world data are in line with the ones observed for the synthetic datasets in Section~\ref{sect:synth-data}. 
Figure \ref{fig:MAB-rw} shows the empirical results obtained for the article whose arms are presented in Figure~\ref{fig:CTRmean}. 
While the empirical pseudo-regret $\hat{R}_T$ seems comparable between the \alg{} algorithms and their traditional counterparts due to the hardness of the problem and the limited horizon, UCBrev+ clearly shows an higher regret (Figure~\ref{fig:MAB-rw}(a)).
Indeed, Figure~\ref{fig:MAB-rw}(b) shows that even though the value of the $NPR$ of \alg(bUCB) and \alg(TS) decreases during the simulations, it does not converge yet to $1$, as occurred for the synthetic dataset (Figure~\ref{fig:MAB-synth10}), meaning that the algorithms are still pulling multiple arms per round.
However, both algorithms show significantly higher $Opti^*$ and marginally higher $Opt^*$ compared to their respective counterparts indicating a better selection of the optimal article during the simulations.

\begin{figure}[t!]
	\centering
	\includegraphics[width=0.75\textwidth]{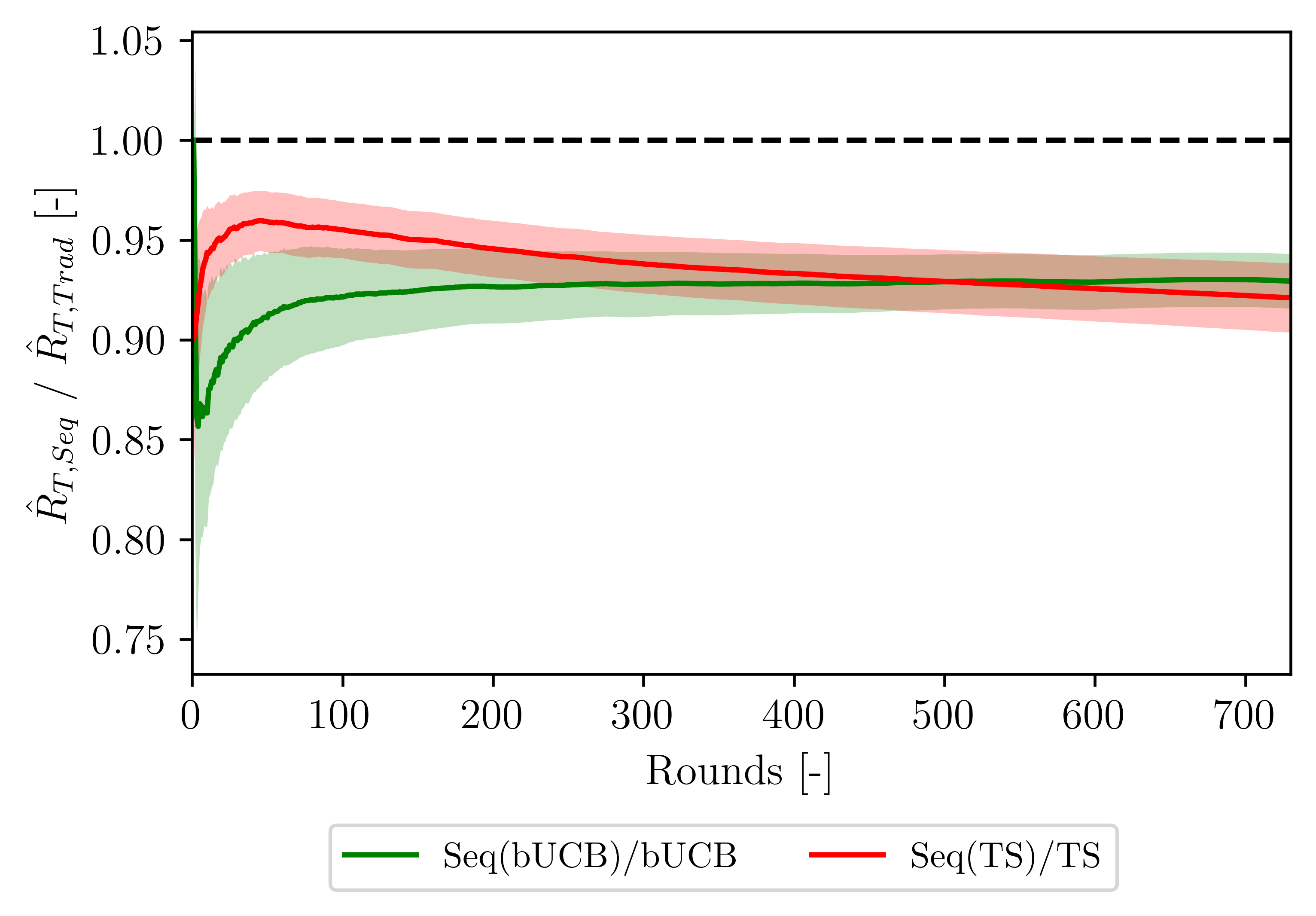}
	\caption{Ratio of $\hat{R}_T$ of the \alg{} algorithms and their traditional counterparts.}
	\label{fig:RegretRatio-rw}
\end{figure}

To evaluate the performance of the \alg{} algorithms and their counterparts against all the article placement problems and to better highlight the difference in the empirical pseudo-regret of the two algorithms pairs, the ratio between the empirical pseudo-regret of \alg{} algorithms and their traditional counterparts was calculated for all the articles.
A value of this ration smaller than $1$ provides evidence for the sequential counterpart provides a smaller regret.
Such a measure allows to compare directly the difference due to the adoption of \alg{} approach regardless of the different number of arms and hardness of the different problems.
Figure~\ref{fig:RegretRatio-rw} shows that the mean and its confidence interval of the ratios of the $Seq(bUCB)$ and $Seq(TS)$ algorithms are below $1$ over the entire time horizon $T$.
Even if $Seq(bUCB)$ seems to perform better during an initial period, the two algorithms converge to a rate of $\approx 0.9$ by the end of the time horizon.
Such result indicate how both algorithms, on average, lead to a $10\%$ reduction of the $\hat{R}_T$ with respect to their traditional counterparts.

\subsubsection{Best-Arm Identification}

The performance in a BAI settings were tested using a dataset regarding environmental monitoring.
In this case-study the identification of maximum pollutants concentrations within a streaming environment (i.e., drinking water distribution system, surface water) is critical to properly estimate the health or environmental risks connected to their presence in the given environment.
The data, described in detail in~\cite{Gabrielli2021env}, consist of bacterial concentrations in a drinking water distribution system monitored every $2$ hours, i.e., we have $K = 12$.
The goal is to detect the hour of the day, among the available ones, for which the bacterial concentrations overcomes a warning threshold $\Gamma = 60~\frac{cells}{\mu L}$ with the largest probability.
At the same time, since the measurements are expensive, we would like to limit the number of measurements performed to perform the detection task.
The samples collected over time are Bernoulli realizations of the measurement, stating if at a specific time the threshold $\Gamma$ has been exceeded or not.
We analyzed the data corresponding to this phenomenon over a period of $26$ days, during which the data can be considered stationary over time, and estimated the probabilities of exceeding the threshold $\Gamma$ for each measurement hour, which are used as values for the expected reward $\mu_i$ in the evaluation.
To provide a sufficiently long time horizon for the MAB algorithms, the selected stationary period was repeated until reaching a total of $494$ days, i.e., rounds.

\begin{figure}[t!]
	\centering
	\includegraphics[width=\textwidth]{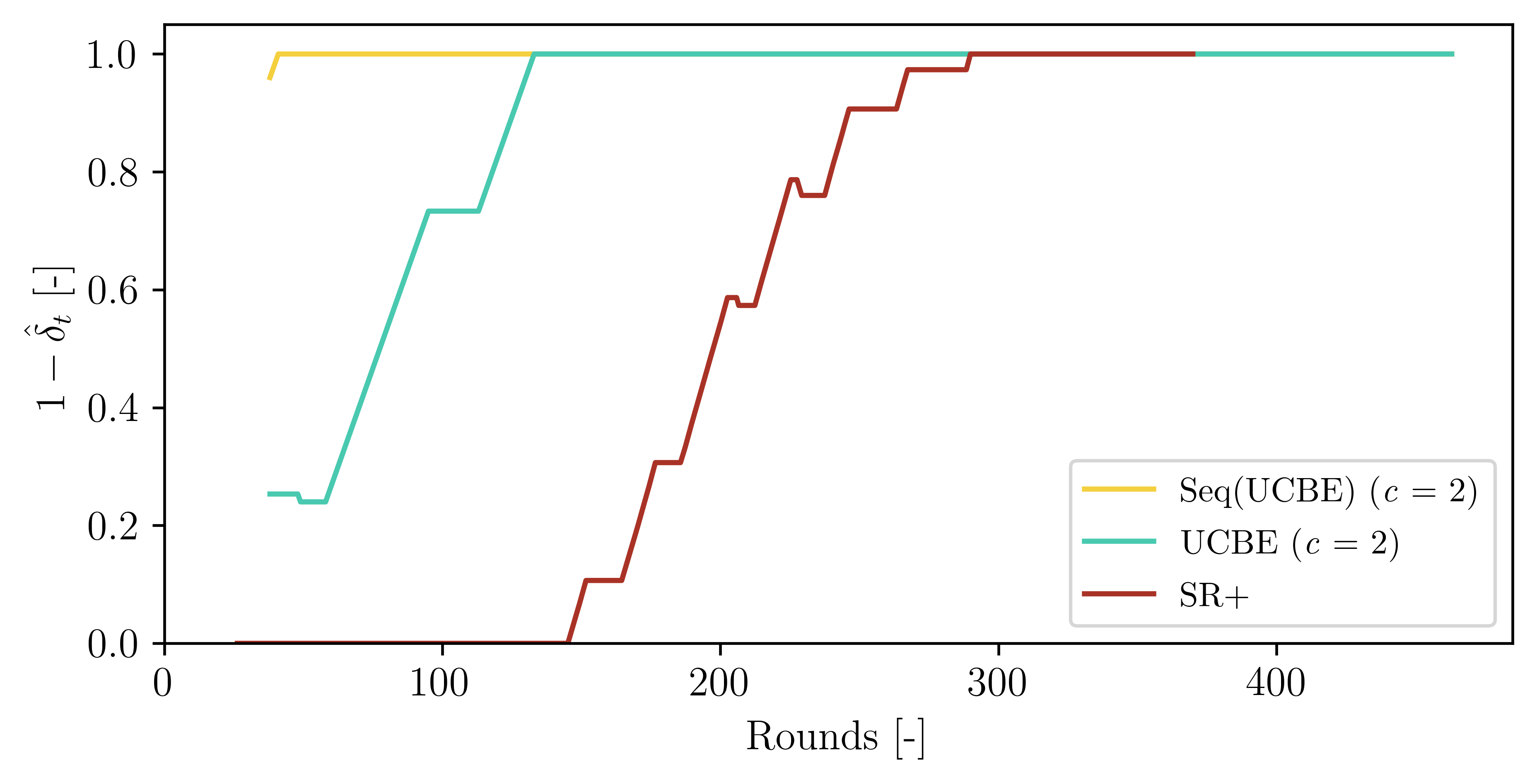}
	\caption{Percentage of correct best-arm identifications ($1 - \hat{\delta_t}$) (rolling mean, $n = 75$) with respect to the number of rounds used in the real-world problem ($\Gamma = 60~\frac{cells}{\mu L}$) for Seq(UCBE) ($c = 2$), UCBE ($c = 2$) and SR+.}
	\label{fig:BAI-rw-c2}
\end{figure}

The environmental monitoring data was used to test the proposed approach in a BAI context.
Figure~\ref{fig:BAI-rw-c2} shows that Seq(UCBE) requires a lower number of rounds to identifying the best arm correctly than both UCBE and SR+.\footnote{The results corresponding to other values of $c$ are reported in Appendix~\ref{app:additional}, and are in line with those presented in this section.}
Notice that a faster correct identification enables a more prompt intervention to remove the cause of increased concentrations and to reduce the cost linked with the monitoring campaign.
Therefore, the adoption of the proposed technique is of paramount importance for this application.

	\section{Conclusions and Future Works}

This paper has formalized a novel MAB setting, namely {Sequential Pull/No-pull Bandit}, which includes a temporal dependency in how the arms are presented to the learner.
We proposed a meta-algorithm, namely \alg, to adapt any classical MAB algorithms, either for RM or BAI, and revised two state-of-the-art elimination algorithms to better suit the proposed setting.
The proposed meta-algorithm provides the same theoretical guarantees as the classical MAB algorithm employed for both RM and BAI problems and, depending on the classical policy used, shows stronger empirical performances on both synthetic and real-world datasets.

Interesting research lines are the extension of the proposed meta-algorithm to a non-stationary environment, to exploit explicitly the correlation existing among the available arms, and the development of \textit{ad hoc} algorithms with stronger theoretical guarantees.

	\bibliographystyle{abbrv}
	\bibliography{biblio}
	
	\clearpage
	\appendix
	\section*{Supplementary Material}
\renewcommand\thefigure{S\arabic{figure}}   
\setcounter{figure}{0}
\renewcommand\thealgorithm{S\arabic{algorithm}}   
\setcounter{algorithm}{0}

\section{Proofs} \label{app:proof}

\thmregclassical*

\begin{proof}
	In the SPNB setting the number of times we are allowed to make a decision is equal to the number of phases $\tau$, and, therefore, the corresponding classical MAB setting has this time horizon.
	Since the regret can be rewritten as follows:
	\begin{equation} \label{eq:reg1}
		R_\tau(\mathfrak{U}) = \sum_ {a_i \neq a^*} L_i \mathbb{E}[T_i(\tau)],
	\end{equation}
	where $L_i$ is the expected loss (w.r.t.~pulling the optimal arm for the current phase) we incur by pulling an arm and $\mathbb{E}[T_i(\tau)] \leq C \log(\tau) + A$ is the expected number of times the arm $a_i$ is pulled over $\tau$ rounds.
	In the SPNB setting each time we pull a suboptimal, besides getting an expected negative reward of $\mathbb{E}[X_{*,t} - X_{i,t}] = \Delta_i$, we also prevent the algorithm to pull the optimal one, getting an expected loss of $\mathbb{E}[X_{*,t}] = \mu^*$, therefore, we suffer a loss of $L_i = (\mu^* + \Delta_i)$.
	Substituting the expression of the expected loss and expected number of pulls and using the fact that $T = k \tau$ in the regret provides the final result.
\end{proof}

\thmregseq*

\begin{proof}
	Restricting the analysis to the time instant during which we pulled an arm we are following the given MAB.
	Denote with $T'$ the number of pulls at the end of the time horizon $T$ which satisfies $\tau \leq T' \leq T$ since at least one pull per round should have been performed and at most $K$.
	Therefore, we can perform the analysis on an effective time horizon of $T'$.
	The loss in a round is equal to $\Delta_i$ if the arm is suboptimal and we pull it, and equal to $\mu^*$ if the arm is the optimal one and we do not pull it.
	The number of times we pull a suboptimal arm $a_i$ is bounded by $C_i \log(T') + A_i$ over $T'$ rounds.
	Conversely, the number of times we do not pull the optimal arm can be written as:
	\begin{align*}
		& \sum_{t=1}^{T'} \mathbb{P}(a_t \neq a^*) = \sum_{t=1}^{T'} \mathbb{P}(\exists a_i \neq a^* \wedge a_i = a_t) \leq \sum_{t=1}^{T'} \sum_{a_i \neq a^*} \mathbb{P}(a_i = a_t) \\
		&= \sum_{a_i \neq a^*} \underbrace{\sum_{t=1}^{T'} \mathbb{P}(a_i = a_t)}_{\mathbb{E}[T_i(T')]} \leq \sum_{a_i \neq a^*} [C_i \log(T') + A_i].
	\end{align*}
	Overall, the regret becomes:
	\begin{align*}
		&R_T(Seq(\mathfrak{U}_{MAB})) \leq \sum_{a_i \neq a^*} \Delta_i [C_i \log(T') + A_i] + \mu^* \sum_{a_i \neq a^*} [C_i \log(T') + A_i] \\
		&\leq \sum_{a_i \neq a^*} (\Delta_i+\mu^*) [C_i \log(K \tau) + A_i] = \sum_{a_i \neq a^*} (\Delta_i+\mu^*) [C_i \log(\tau) + A'_i],
	\end{align*}
	where we defined $A'_i := (\Delta_i+\mu^*) [C_i \log(K) + A_i]$ and we used the fact that $T' \leq T = K \tau$.
\end{proof}

\thmbaisr*

\begin{proof}
	The SR algorithm pulls the last two arms a number of times equal to $\lceil \frac{1}{\overline{\log}(K)} \frac{n-K}{2}\rceil$, where $n$ is the total number of pulls available as budget.
	Setting $n = \frac{(2 T - 1) \overline{\log}(K)}{2K}+K$ would require to get a number of rounds:
	\begin{align}
		\left\lceil \frac{1}{\overline{\log}(K)} \frac{n-K}{2} \right\rceil = \left\lceil \frac{2T-1}{2K} \right\rceil \leq \tau,
	\end{align}
	which is compatible with the SPNB framework.
	Substituting the expression of $n$ in the result of Theorem $2$ of~\cite{audibert2010best} provides the final result.
\end{proof}

\begin{restatable}[]{mythm}{thmregucbrev} \label{thm:regucbrev}
	Using the UCBrev+ algorithm for a SPNB setting, it suffers a pseudo-regret of:
	\begin{align}
		R_T(UCBrev+) & \leq \sum_{a_i \in A \ | \ \Delta_i > \lambda} \left( \Delta_i + \frac{32 \log(T \Delta_i^2)}{\Delta_i} + \frac{96}{\Delta_i} + \frac{512 \mu^*}{\Delta_i^2} \right) \nonumber \\ 
		& + \sum_{a_i \in A \ | \ 0 \leq \Delta_i \leq \lambda} \left( \frac{64}{\lambda} + \frac{512 \mu^*}{\lambda^2} + \Delta_i T \right),
	\end{align}
	for each $\lambda \geq \sqrt{\frac{e}{T}}$.
\end{restatable}

\begin{proof}
	The proof follows the steps of Theorem $3.1$ in~\cite{Auer2010ucbrev}, where the definition of regret changes according to what has been defined for the SPNB setting.

	Recall that the algorithm works in phases $m \in \{0, \ldots, \lfloor \frac{1}{2} \log_2 \frac{\tau}{e} \rfloor \}$, and the proof decompose the regret suffered in each phase.
	In the SPNB setting this algorithm continues to run until it finishes all the $\tau$ rounds, using a total of $T' \leq T$ pulls.

	At first, the proof focus on those arms having a gap $\Delta_i > \lambda$ for some fixed $\lambda \geq \sqrt{\frac{e}{\lambda}}$, where the other arms will provide a regret of $\sum_{a_i | \Delta_i \leq \lambda} \Delta_i T' \leq \sum_{a_i | \Delta_i \leq \lambda} \Delta_i T$.
	
	After that, the proof divides the regret of the other arms into $3$ contributions:
	\begin{itemize}
		\item $R_a$: some suboptimal arm $a_i$ is not eliminated in round $m_i := \min\{m \ | \ \hat{\Delta}_m \leq \frac{\Delta_i}{2} \}$ (or before), still having the optimal arm in the set of available ones;
		\item $R_b$: each suboptimal arm $a_i$ has been eliminated in round $m_i$ (or before);
		\item $R_c$: the optimal arm $a^*$ is eliminated by some suboptimal arm $a_i$ in at round $m^*$.
	\end{itemize}

	Let us define $A' := \{a_i | \Delta_i \geq \lambda\}$ and $A' := \{a_i | \Delta_i > 0\}$.
	The contribution to the regret $R_a$ is the same as in the standard MAB setting:
	\begin{equation}
		R_a \leq \sum_{a_i \in A'} \frac{32}{\Delta_i}.
	\end{equation}
	
	Using the definition of $m_i$, the regret of $R_b$ is bounded as:
	\begin{equation}
		R_b \leq \left( \Delta_i + \frac{32 \log(T' \Delta_i^2)}{\Delta_i} \right) \leq \left( \Delta_i + \frac{32 \log(T \Delta_i^2)}{\Delta_i} \right),
	\end{equation}
	where the last inequality is from the fact that the effective number of pulls of the algorithm $T'$ is smaller or equal than the time horizon $T$.
	
	Instead, the contribution of $R_c$ is different from the classical MAB setting, since the fact that we eliminated an arm implies that the regret per round has an additional $\mu^*$ term.
	This leads to the following:
	\begin{align}
		R_c &\leq \sum_{m^* = 0}^{\max_{j \in A'} m_j} \sum_{a_i \in A'' | m_i \geq m^*} \frac{2}{T \hat{\Delta}^2_{m^*}} T \left(\max_{j \in A'' | m_j \geq m^*} \Delta_j + \mu^* \right)\\
		&\leq \sum_{a_i \in A'} \left( \frac{64}{\Delta_i} + \frac{512 \mu^*}{\Delta_i^2} \right)+ \sum_{a_i \in A'' \setminus A'} \left( \frac{64}{\lambda} + \frac{512 \mu^*}{\lambda^2} \right).
	\end{align}
	
	Adding the three components of the regret to the contribution of regret given from arms in $A'' \setminus A'$ provides the final statement of the theorem.
\end{proof}

Notice that this proof is able to reduce the multiplicative constant of the $O(\log(T))$ term of a factor $\frac{\Delta_i + \mu^*}{\Delta_i} = 1 + \frac{\mu^*}{\Delta_i}$ w.r.t.~the one present in Theorem~\ref{thm:classical}.
This comes at the cost of using the UCBrev algorithms, whose constant is not optimal for this problem.
Therefore the potential improvement provided by the fact that in high probability the optimal arm is never discarded during the process, is overcome by the intrinsic worse guarantees of the original UCBrev algorithm.

\section{Algorithms} \label{app:algo}

In what follows we present the pseudo-code of the algorithms used to apply standard RM and BAI algorithm to the SPNB setting. \\

A generic $\mathfrak{U}_{MAB}$ algorithm can be applied to the SPNB setting by splitting the entire time horizon $T$ in a sequence of $\tau$ rounds, where at the onset of each round the learner can select a single arm. After pulling the selected arm and collecting the feedback, the learner updates its arms estimates and proceeds to the next round, as shown in Algorithm \ref{alg:mab_generic}.\\

\begin{algorithm}
	\caption{$\mathfrak{U}_{MAB}$}
	\label{alg:mab_generic}
	\begin{algorithmic}[1]
		\State \textbf{Input:} arm set $\{ a_1, \ldots, a_K \}$, time horizon $T$
		\State Initialize $\mathfrak{U}_{MAB}$
		\State $n \gets  0$
		\State $\tau \gets \frac{T}{K}$
		\State $a_{MAB} \gets \mathfrak{U}_{MAB}(n)$
		\For{$t \in \{1, \ldots, \tau\}$}
		\State Pull arm $a_{MAB}$
		\State Collect feedback $x_{MAB,t}$
		\State $n \gets n + 1$
		\State Update $\mathfrak{U}_{MAB}$
		\State $a_{MAB} \gets \mathfrak{U}_{MAB}(n)$
		\EndFor
	\end{algorithmic}
\end{algorithm}

The UCBrev+ algorithm (Algorithm \ref{alg:ucbrev}) can be employed to the SPNB setting by dividing the time horizon $T$ in $\tau$ rounds and pulling all the available arms in each round. At specified rounds, and if more than one arm is still available, the average reward of each available arm is computed and the arms whose upper bound is smaller than the maximum lower bound are dismissed, as done in the traditional version. \\ 

\begin{algorithm}
	\caption{$UCBrev+$}
	\label{alg:ucbrev}
	\begin{algorithmic}[1]
		\State \textbf{Input:} arm set $\{ a_1, \ldots, a_K \}$, time horizon $T$
		\State Initialize $UCBrev+$
		\State $\tau \gets \frac{T}{K}$
		\State $\tilde{\Delta}_0 \gets  1$
		\State $B_0 \gets \{ a_1, \ldots, a_K \}$
		\State $m \gets 0$
		\For{$t \in \{1, \ldots, \tau\}$}
		\If{$t = \lceil \frac{2log(\tau\tilde{\Delta_m}^2)}{\tilde{\Delta}_m^2} \rceil $ \textbf{and} $ |B_m| > 1$}
		\State $m \gets m + 1$
		\State Compute arms average reward $\mu_i ~\forall i \in B_{m-1}$
		\State $B_m \gets B_{m-1} \setminus \left\lbrace\left\lbrace  \mu_i + \sqrt{\frac{\log(t\tilde{\Delta}_{m-1}^2)}{2t}}\right\rbrace_{\forall i \in B_{m-1}}  < max_{j \in B_{m-1}}\left\lbrace  \mu_j - \sqrt{\frac{\log(t\tilde{\Delta}_{m-1}^2)}{2t}}\right\rbrace\right\rbrace$
		\State $\tilde{\Delta}_m \gets  \frac{\tilde{\Delta}_{m-1}}{2}$
		\EndIf
		\State Pull all arms $\in B_{m}$
		\EndFor
	\end{algorithmic}
\end{algorithm}

The SR+ algorithm (Algorithm \ref{alg:sr}) adapts the traditional version of SR to the SPNB setting by splitting the time horizon $T$ in $\tau$ rounds, which are, then splitted, in $K - 1$ phases. All the available arms are pulled at each round. At the end of each phase the arm with the lowest empirical mean $\hat{\mu}_i$ is dismissed. The recommended arm is the last, and single, available arm present.

\begin{algorithm}
	\caption{$SR+$}
	\label{alg:sr}
	\begin{algorithmic}[1]
		\State \textbf{Input:} arm set $\{ a_1, \ldots, a_K \}$, time horizon $T$
		\State Initialize $SR+$
		\State $\tau \gets \frac{T}{K}$
		\State $k \gets \{1, \ldots, K-1\}$
		\State $\overline{log}(K) = \frac{1}{2} + \sum_{i=2}^{K}\frac{1}{i}$
		\State $n = \left\lbrace \lceil \frac{1}{\overline{log}(K)}\frac{\tau	-K}{K+1-i}\rceil \right\rbrace_{\forall i \in k}$
		\State $B_0 \gets \{a_1, \ldots, a_K\}$
		\State $k \gets 0$
		\For{$t \in \{1, \ldots, \tau\}$}
		\If{$t = n_k$}
		\State $k \gets k + 1$
		\State Compute arms average reward $\mu_i ~\forall i \in B_{k-1}$
		\State $B_k \gets B_{k-1} \setminus min_{i \in B_{k-1}}~\hat{\mu}_{i}$
		\EndIf
		\State Pull all arms $\in B_{k}$
		\EndFor
		\State \textbf{Output:} $\hat{a}^* = B_k$
	\end{algorithmic}
\end{algorithm}

\clearpage

\section{Supplementary Experimental Results}

\subsection{BAI arms distribution}
The distribution of the feedback $\mu_i$ adopted in the synthetic datasets used to evaluate the performance of the algorithms in a BAI settings follows the experiments proposed by Audibert and coworkers~\cite{audibert2010best}. Bernoulli distributions were considered for the arms in all experiments. While the optimal arm was assigned $\mu_1 = 0.5$, the number of arms and their $\mu_i$ distribution varied between experiments as shown below, where $K$ represents the number of arms:
\begin{itemize}
	\item Experiment 1: $K = 20, \mu_{2:20} = 0.4$;
	\item Experiment 2: $K = 20, \mu_{2:6} = 0.42, \mu_{7:20} = 0.38$;
	\item Experiment 3: $K = 4, \mu_i = 0.5 - (0.37)^i, i \in {2,3,4}$;
	\item Experiment 4: $K = 6, \mu_2 = 0.42, \mu_{3:4} = 0.4, \mu_{5:6} = 0.35$;
	\item Experiment 5: $K = 15, \mu_I = 0.5 - 0.025i, i \in {2,\dots,15}$;
	\item Experiment 6: $K = 20, \mu_2 = 0.48, \mu_{3:20} = 0.37$;
	\item Experiment 7: $K = 30, \mu_{2:6} = 0.45, \mu_{7:20} = 0.43, \mu_{21:30} = 0.38$.
\end{itemize}

\subsection{Additional Experimental Results} \label{app:additional}
In what follows we provide the results of the synthetic datasets corresponding to $K = 10$ and $K = 50$ for the RM setting and the ones corresponding to $c \in \{1, 4, 8\}$ for the BAI setting, and the BAI results obtained with $c \in \{1, 2, 4, 8\}$ in the real-world dataset. \\

\FloatBarrier

\begin{figure}[t!]
	\centering
	\includegraphics[width=\textwidth]{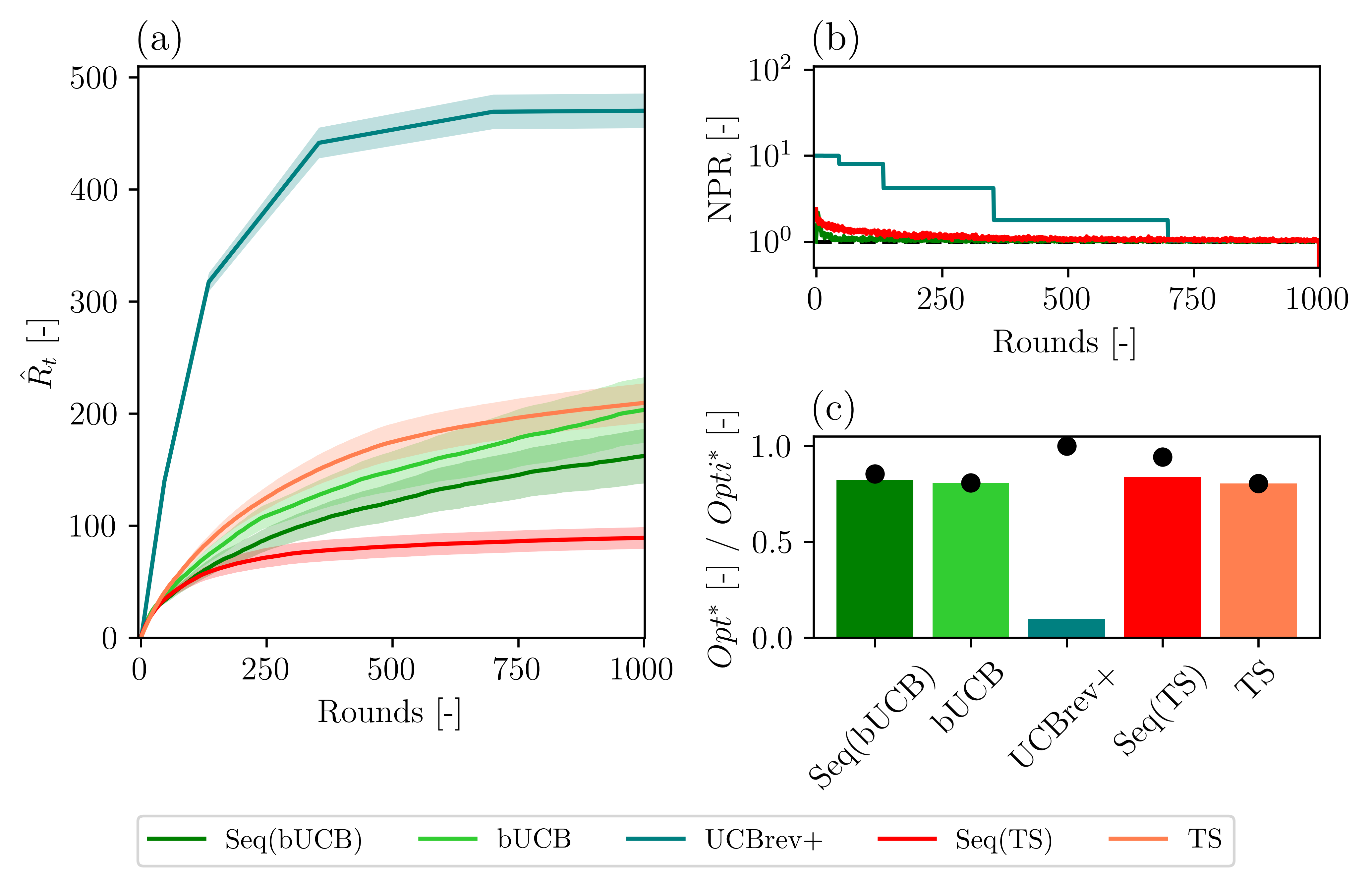}
	\caption{Results for the synthetic setting with $K = 10$ arms: (a) $\hat{R}_T$, (b) NPR , (c) $Opt^*$ (shown as bars) $Opti^*$ (shown as points) for the analyzed algorithms.}
	\label{fig:MAB-synth10}
\end{figure}

\begin{figure}[t!]
	\centering
	\includegraphics[width=\textwidth]{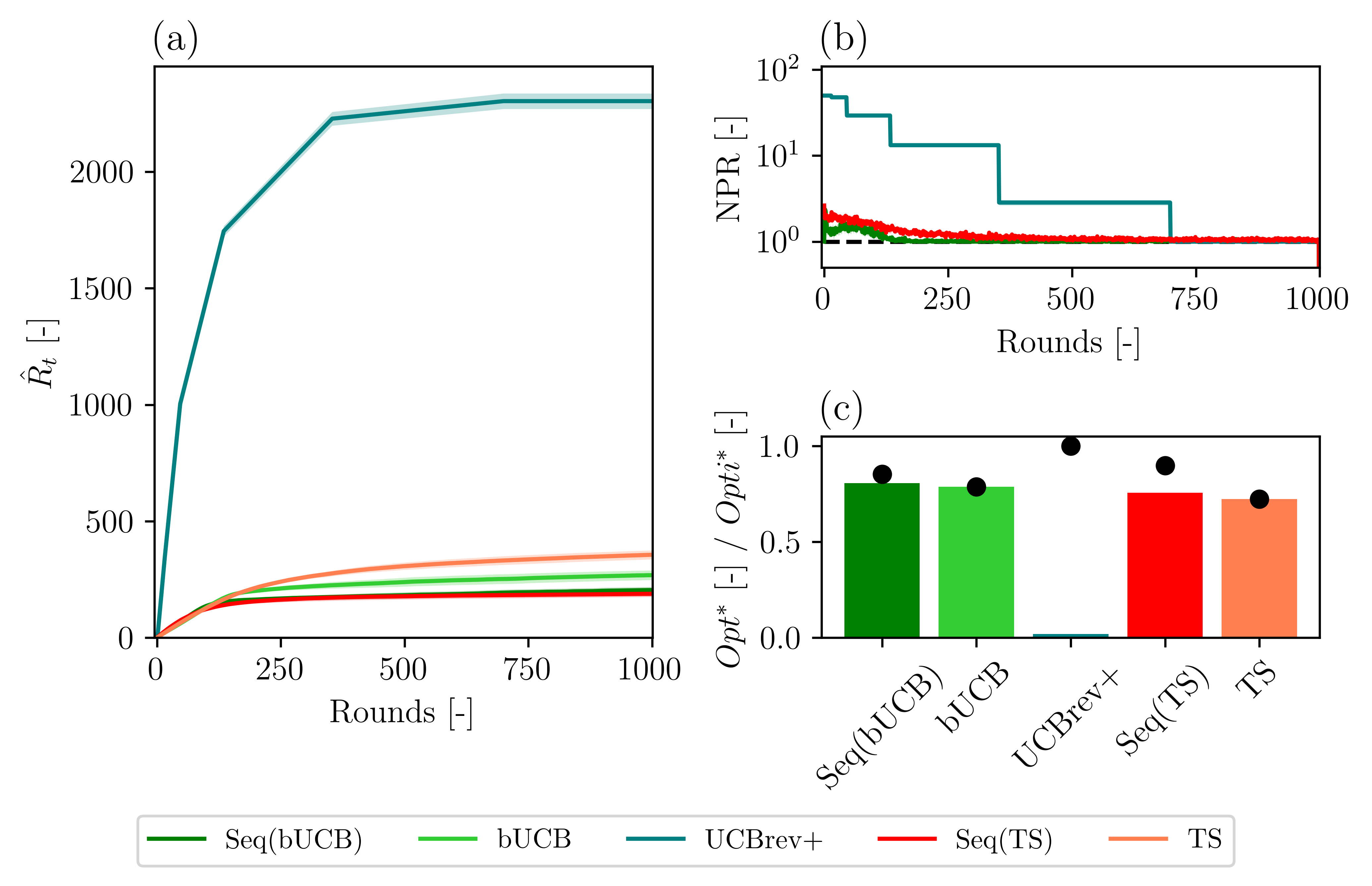}
	\caption{Results for the synthetic setting with $K = 50$ arms: (a) $\hat{R}_T$, (b) NPR , (c) $Opt^*$ (shown as bars) $Opti^*$ (shown as points) for the analyzed algorithms.}
	\label{fig:MAB-synth50}
\end{figure}

While overall similar results are obtained with a different $K$ in the synthetic RM experiments highlighting how the proposed Seq approach provides better performances compared to the use of traditional algorithms regardless of the number of arms of the problem (Figures~\ref{fig:MAB-synth10} and~\ref{fig:MAB-synth50}).\\

\begin{figure}[t!]
	\centering
	\includegraphics[width=\textwidth]{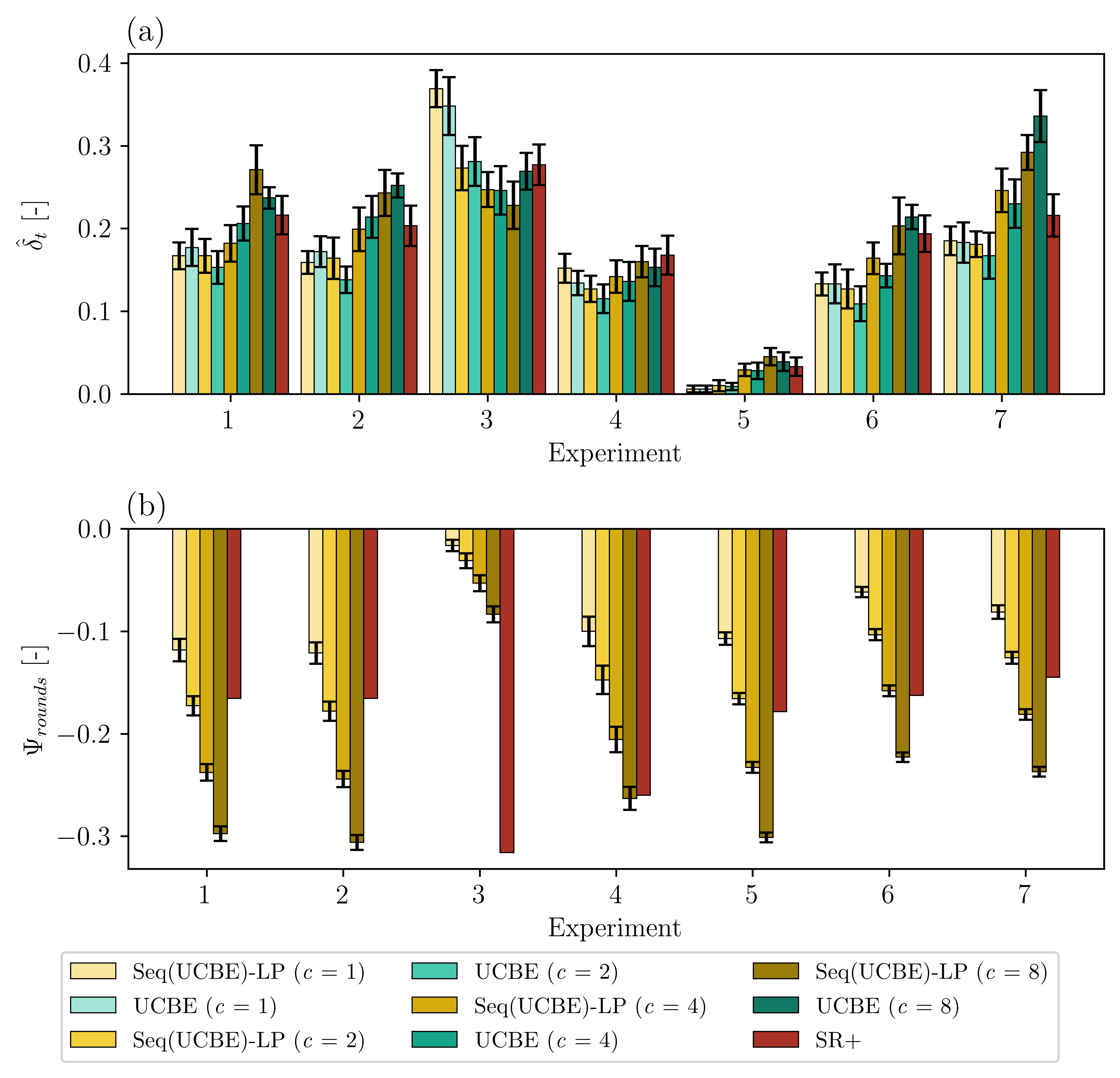}
	\caption{Probability of selecting a suboptimal arm as best (a), percentage of rounds used before selecting the best arm with respect to the UCBE algorithm (b) for for the analyzed algorithms.}
	\label{fig:BAI-synth-LP}
\end{figure}

\begin{figure}[t!]
	\centering
	\includegraphics[width=\textwidth]{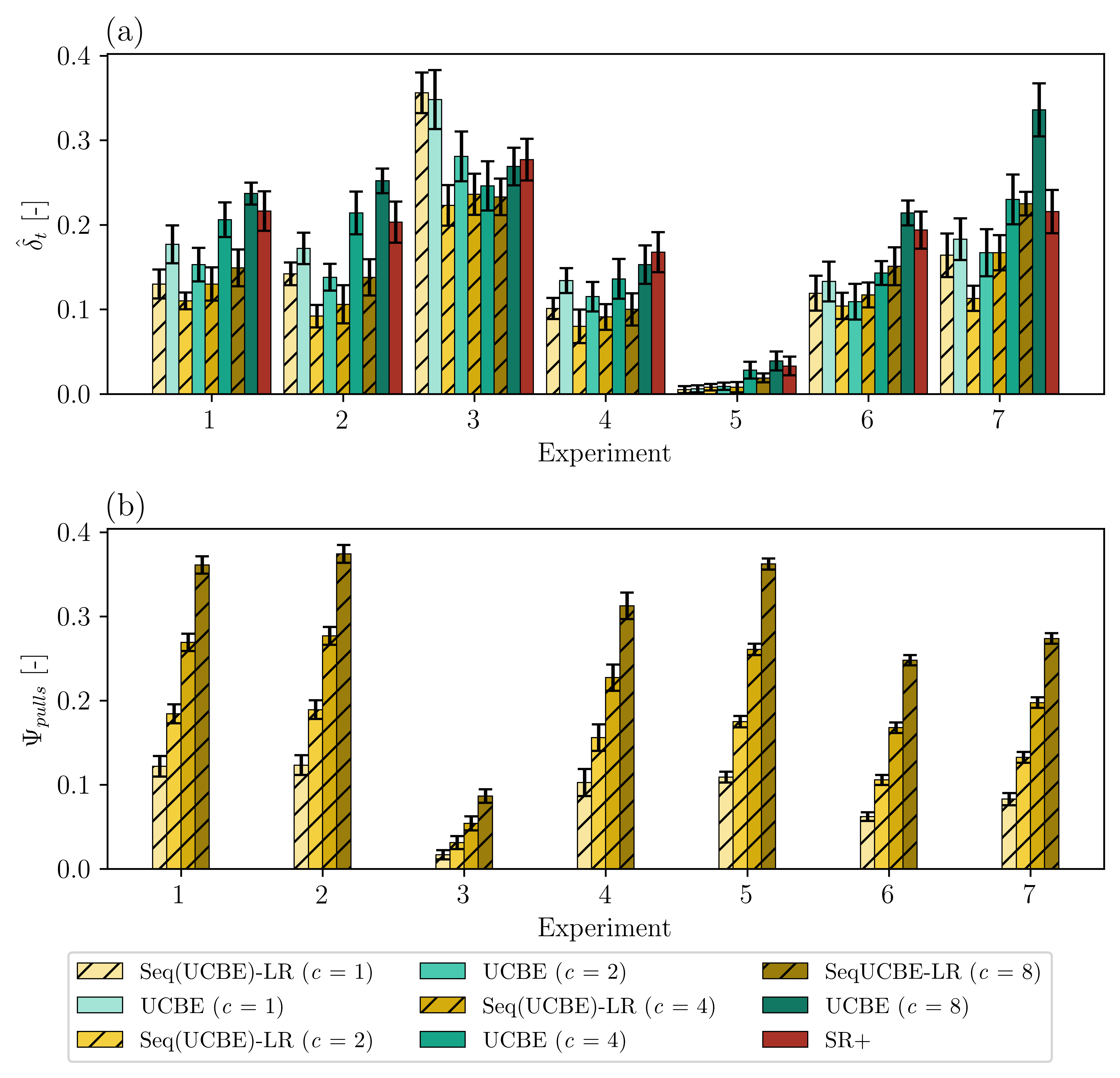}
	\caption{Probability of selecting a suboptimal arm as best (a), percentage of pulls used before selecting the best arm with respect to UCBE and SR+ algorithms (c) for the analyzed algorithms.}
	\label{fig:BAI-synth-LR}
\end{figure}

By observing the complete BAI simulation results for the synthetic dataset, it can be observed how $c$, other than varying UCBE and Seq(UCBE) performances, influences monotonically both the reduction of rounds of Seq(UCBE)-LP and the increase percentage of pulls of Seq(UCBE)-LR (Figures~\ref{fig:BAI-synth-LP} and~\ref{fig:BAI-synth-LR}).\\

\begin{figure}[t!]
	\centering
	\includegraphics[width=\textwidth]{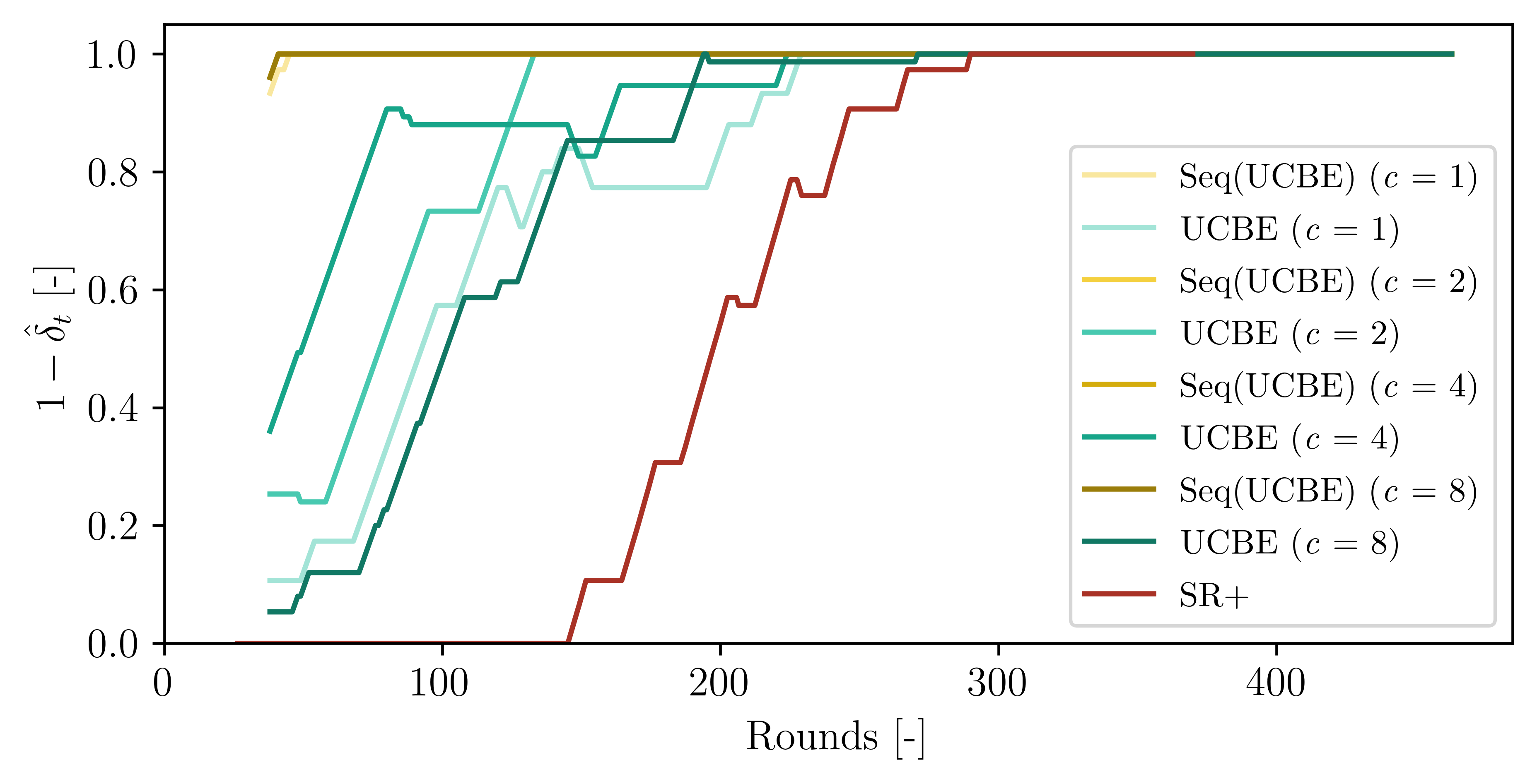}
	\caption{Percentage of correct best-arm identifications ($1 - \hat{\delta_t}$) (rolling mean, $n = 75$) with respect to the number of rounds used in the real-world problem ($\Gamma = 60~\frac{cells}{\mu L}$).}
	\label{fig:BAI-rw-compl}
\end{figure}

As observed for $c = 2$, for every value of $c$ tested Seq(UCBE) yields higher percentages of correct best-arm identification ($1 - \hat{\delta_t}$) than both UCBE and SR+ (Figure~\ref{fig:BAI-rw-compl}).

\section{Technical details}
The experiments were carried on a PC running Windows $10$ and equipped with a Intel i7-4790 processor and $8$ Gb of RAM.\\
The libraries used are:
\begin{itemize}
	\item Python 3.7.9
	\item Numpy 1.19.1
	\item Pickles 0.0.11
	\item Matplotlib 3.3.1
\end{itemize}
The average running time for the synthetic MAB experiments was of $18$ s, $43$ s, and $1$ min $42$ s (on average) for the settings with $10$, $25$ and $50$ arms, respectively.

\end{document}